\documentclass[letterpaper, 10 pt, conference]{ieeeconf}  

\usepackage{amsfonts}
\usepackage{xspace}
\usepackage{xcolor}
\usepackage{graphicx}
\usepackage{amsmath}
\usepackage{tabularx}
\usepackage{booktabs}
\usepackage{silence}
\usepackage[font=normal,compatibility=false]{caption}
\usepackage{multirow}
\usepackage{hyperref}
\usepackage[most]{tcolorbox}

\DeclareMathOperator*{\argmin}{argmin}
\newcommand{\algo}{Dream2Flow\xspace}
\newcommand{\flow}{3D object flow\xspace}

\tcbset{
  colback=gray!5!white,
  colframe=gray!50!black,
  boxrule=0.5pt,
  arc=2mm,
  left=1mm,
  right=1mm,
  top=1mm,
  bottom=1mm,
  fonttitle=\bfseries,
}

\usepackage[noadjust]{cite}

\newtcolorbox{promptbox}[1][]{%
  title=Prompt,
  #1
}

\usepackage{placeins}

\title{\LARGE \bf
\algo: Bridging Video Generation and\\Open-World Manipulation with 3D Object Flow
}

\author{
Karthik Dharmarajan, Wenlong Huang, Jiajun Wu, Li Fei-Fei*, Ruohan Zhang* \\
Stanford University
}

\begin{document}

\makeatletter
\let\@oldmaketitle\@maketitle
\renewcommand{\@maketitle}{\@oldmaketitle
    \vspace{0.15cm}
    \begin{center}
    \captionsetup{hypcap=false}
    \includegraphics[width=0.99\linewidth]{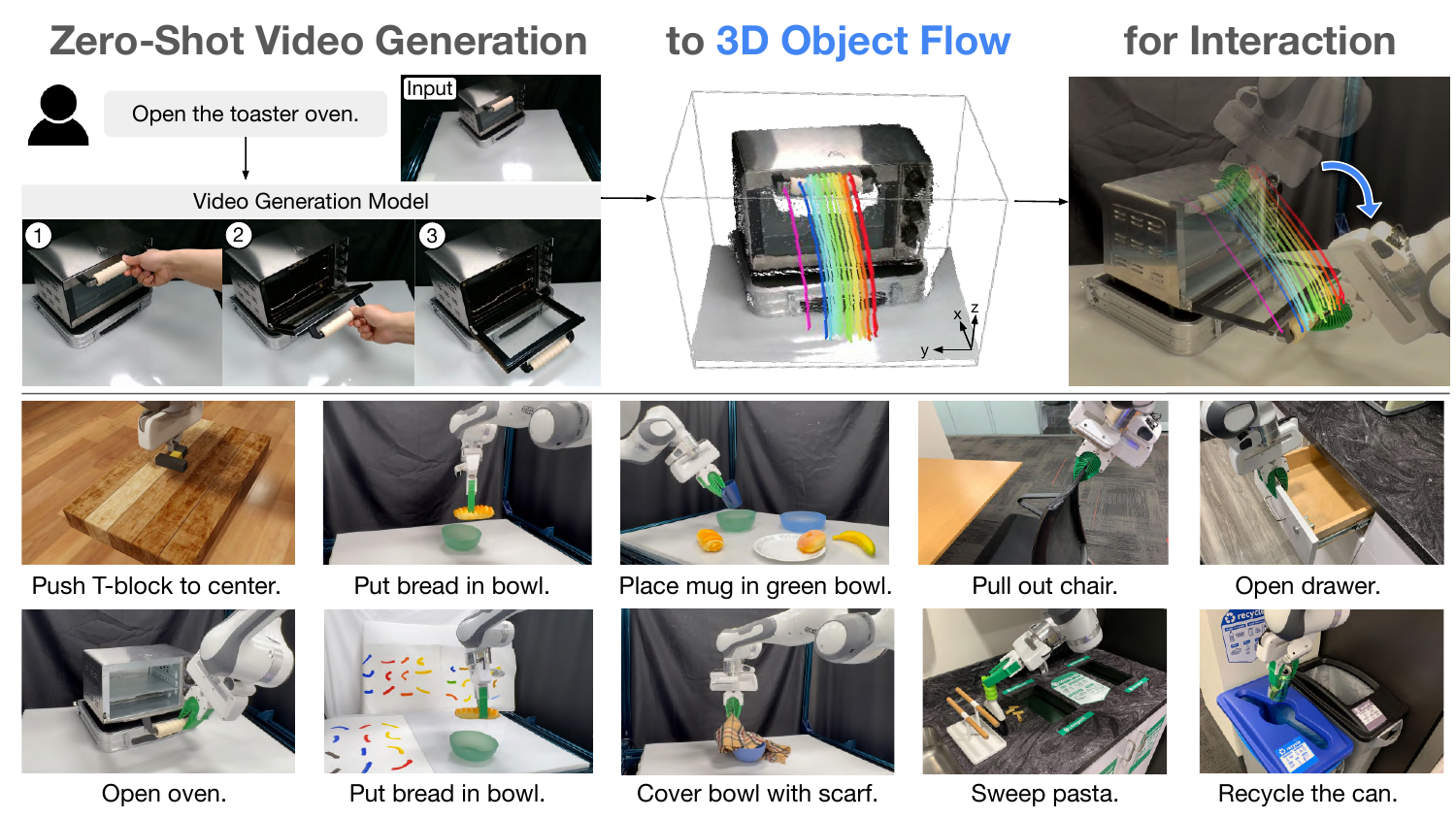}
    \captionof{figure}{\textbf{\algo} leverages off-the-shelf video generation models to produce videos of the task being performed in the same scene of the robot. \algo then extracts a 3D object flow from the motion in the video, allowing for downstream planning and execution with a robot across a wide variety of tasks.}
    \label{fig:pull_fig}
    \vspace{-0.35cm}
    \end{center}}
\makeatother

\maketitle
\addtocounter{figure}{-1}

\thispagestyle{empty}
\pagestyle{empty}

\begin{abstract}
Generative video modeling has emerged as a compelling tool to zero-shot reason about plausible physical interactions for open-world manipulation. Yet, it remains a challenge to translate such human-led motions into the low-level actions demanded by robotic systems. We observe that given an initial image and task instruction, these models excel at synthesizing sensible object motions. Thus, we introduce \textbf{\algo}, a framework that bridges video generation and robotic control through \textbf{\flow} as an intermediate representation. Our method reconstructs 3D object motions from generated videos 
and formulates manipulation as object trajectory tracking. 
By separating the state changes from the actuators that realize those changes, \algo overcomes the embodiment gap and enables zero-shot guidance from pre-trained video models to manipulate objects of diverse categories---including rigid, articulated, deformable, and granular. Through trajectory optimization or reinforcement learning, \algo converts reconstructed \flow into executable low-level commands without task-specific demonstrations. 
Simulation and real-world experiments highlight 3D object flow as a general and scalable interface for adapting video generation models to open-world robotic manipulation. Videos and visualizations are available at \href{https://dream2flow.github.io/}{https://dream2flow.github.io/}.
\end{abstract}

\section{Introduction}
\begingroup
  \setlength{\skip\footins}{0pt}    
  \setlength{\footnotesep}{0pt}     
  \let\thefootnote\relax
  \begin{NoHyper}
  \footnotetext{
    *Equal Advising. Correspondence:
    \href{mailto:wenlongh@stanford.edu}{Wenlong Huang}
  }
  \end{NoHyper}
\endgroup
Robotic manipulation in the open world could greatly benefit from visual world models that predict how an environment would evolve given an agent's interactions.
Recent advances in generative video modeling have produced systems capable of zero-shot synthesizing minute-long, high-fidelity clips of physical interactions in pixel space, conditioned on an unseen initial image and an open-ended task instruction~\cite{brooks2024video}.
Such video models implicitly capture intuitive physics and rich priors of object properties and interactions, making them compelling for open-world manipulation settings where a robot is tasked to complete novel tasks in unseen environments through partial observations.

Despite their promise, it remains unclear what role such models should serve in a robot manipulation system. 
Most frontier video generators produce the best interaction clips with a human embodiment. This reflects where supervision is most abundant as human interactions are significantly more broadly documented than those of robots. But this becomes a challenge for using them in robotic manipulation due to the embodiment gap and hence the different action spaces.

We propose extracting actionable signals from their visual predictions of human interactions, which will then be enacted by a robot. 
This essentially separates the state changes in the real world from the actuators that realize those changes. 
Our proposed method, \textbf{\algo}, employs \textbf{\flow} as an intermediate interface that bridges high-level video simulation with low-level robot actions.
\algo works because, despite occasional visual artifacts, state-of-the-art video generation models often predict physically plausible object motions that align with the task intent in open-world manipulation tasks.
Then, given generated videos, instead of trying to directly mimic the human motions for completing a given task, we focus on reconstructing and reproducing the object flows in 3D.

The problem is thus reduced to object trajectory tracking: the robot’s job is to manipulate the object to closely follow the generated flow that the video model imagined. This approach cleanly separates what needs to happen (i.e., state changes in an environment) from how a particular embodiment achieves it with respect to its kinematic and dynamic constraints (i.e., actions). Importantly, it seamlessly interfaces with both motion planners and sensorimotor policies---the extracted object motion in 3D serves as the tracking goal for trajectory optimization or a reinforcement learning policy, which then yields a sequence of low-level robot joint commands.

Leveraging off-the-shelf models and tools, we demonstrate an autonomous pipeline that 1) generates a text-conditioned video of plausible interaction~\cite{brooks2024video}, 2) obtains \flow by performing depth estimation and point tracking using vision foundation models~\cite{ravi2024sam2,karaev2024cotracker3}, and 3) synthesizes robot actions that realize this flow using trajectory optimization and reinforcement learning.
Notably, this design enjoys a number of desirable benefits in manipulation tasks. By first leveraging video models---pre-trained on large corpora of human activities---to interpret and ground open-ended language commands in visual predictions, our system inherits a scalable mechanism for task specification. These predictions are then distilled into reconstructed \flow, a representation that naturally captures diverse object interactions spanning rigid, articulated, deformable, and granular objects. Together, this synergy enables an end-to-end pipeline that performs open-world manipulation directly from visual perception and language, without task-specific data or training.

In summary, our key contributions are:
\begin{itemize}
    \item We propose \flow as an interface for adapting off-the-shelf video generation models for open-world manipulation by formulating it as an object trajectory tracking problem.
    \item We demonstrate its effectiveness by implementing the approach in both simulated and real domains, which performs diverse tasks given only RGB-D observations and language instructions in a zero-shot manner.
    \item We examine the properties of \flow by comparing it with alternative intermediate representations and by studying its key design choices as well as generalization properties.
\end{itemize}

\section{Related Works}
\subsection{Task Specification in Manipulation}

Specifying tasks for the wide range of manipulation problems spans symbolic, learning-based, outcome-driven, and object-centric interfaces. Classical approaches encode goals and constraints with symbolic formalisms such as PDDL and temporal logics, or optimize cost-augmented formulations \cite{garrett2021integrated,zhao2024survey,toussaint2015logic}. Learning-based systems specify tasks often through language and perception, mapping instructions to actions via language-conditioned visuomotor policies and vision–language–action models \cite{shridhar2022cliport,shridhar2023perceiver,zitkovich2023rt,black2024pi_0,kim2025openvla}. Outcome-based specification sets goals by example observations, e.g., image goals with goal-conditioned policies \cite{lynch2020learning,ebert2018visual,black2023zero,xie2018few,sharma2019third,mendonca2021discovering}, and some works incorporate force targets \cite{adeniji2025feel}. Object-centric alternatives rely on descriptors or keypoints to capture task-relevant structure~\cite{simeonov2022neural}. Recently, foundation models enable higher-level interfaces that compile intent into actionable specifications via code~\cite{liang2023code}, 3D value maps akin to potential fields~\cite{huang2023voxposer}, keypoint relations~\cite{huang2025rekep}, or affordance maps~\cite{tang2025uad}.

\subsection{2D/3D Flow in Robotics}

Dense motion fields—optical flow, point tracks, and 3D scene/object flow—provide an embodiment-agnostic, mid-level interface for manipulation \cite{xu2025flow,yuan2025general}. In scene-centric formulations, policies parameterize or condition on motion in 2D or 3D to decide actions, with point/keypoint interfaces unifying perception and action and with action-flow improving precision \cite{weng2021fabricflownet,goyal2022ifor,seita2023toolflownet,chen2024g3flow,wang2025skil,haldar2025point,guo2025actionsink}. In object-centric formulations, desired object motion is specified independently of embodiment and then converted into actions via policy inference, planning, and optimization \cite{eisner2022flowbot3d,gao2024flip,guo2025flowdreamer,zhi2025flowaction,he2025manitrend,yin2025object}. When actions are absent, retargeting with predicted tracks and dense correspondences offers a practical bridge across embodiments \cite{bharadhwaj2024track2act,ko2023learning,chen2025ecflow}. Advances in perception and tracking such as RGB-D motion-based segmentation, 3D scene flow in point clouds, zero-shot monocular and ToF-based scene flow, rigid-motion learning, refractive flow for transparent objects, deformable and articulated reconstruction, and structured scene representations \cite{shao2018motionseg,liu2018flownet3d,liang2025zeroshotmsf,nvidia2025zeromsf,sander2025tofsceneflow,byravan2016se3nets,tang2023rftrans,duisterhof2024deformgs,kerr2024rsrd,jiang2024roboexp,shorinwa2024splatmover} make the interface reliable. Flow-derived supervision further supports learning, and video generation supplies plausible visual rollouts for planning and imitation \cite{guzey2025bridging,yu2025genflowrl,bharadhwaj2024gen2act,patel2025rigvid}. Our approach follows the object-centric path by reconstructing 3D object flow from language-conditioned generations and tracking it under embodiment constraints, complementing other flow-conditioned policy representations~\cite{weng2021fabricflownet,seita2023toolflownet,eisner2022flowbot3d,gao2024flip,guo2025flowdreamer}.

\subsection{Video Models for Robotics}
Recent work increasingly integrates video models across robotic tasks in various ways \cite{yang2024video,mccarthy2025towards}. They can serve as auxiliary training objectives \cite{wu2023unleashing,seo2022reinforcement,wu2023pre,yang2024spatiotemporal,hu2024video,li2025unified}, as reward models \cite{huang2024diffusion,escontrela2023video,chen2021learning}, as policies \cite{ajay2023compositional,du2023learning}, or as a simulator for the environments \cite{valevski2024diffusion,bruce2024genie}. 
Notably, predictive modeling in robotics can leverage video frame prediction as a form of world model. By simulating future visual observations, these models can enable visual planning and manipulation by anticipating how the environment will evolve. For example, a video generative model can simulate long-horizon task outcomes, effectively acting as a visual planner \cite{luo2024grounding}. Video models in this way can also serve as 
as dynamics models \cite{du2023learning,yang2023learning,zhou2024robodreamer,rybkin2018learning,mendonca2023structured,che2024gamegen}.
Another notable direction is that video generation can directly provide new training data for robot learning. Several recent works devise frameworks to imagine new trajectories in the form of videos and use them to train or finetune policies, particularly for imitation learning \cite{jang2025dreamgen,liang2024dreamitate}.

\section{Method}
\begin{figure*}[!thbp]
    \centering
    \includegraphics[width=0.99\linewidth]{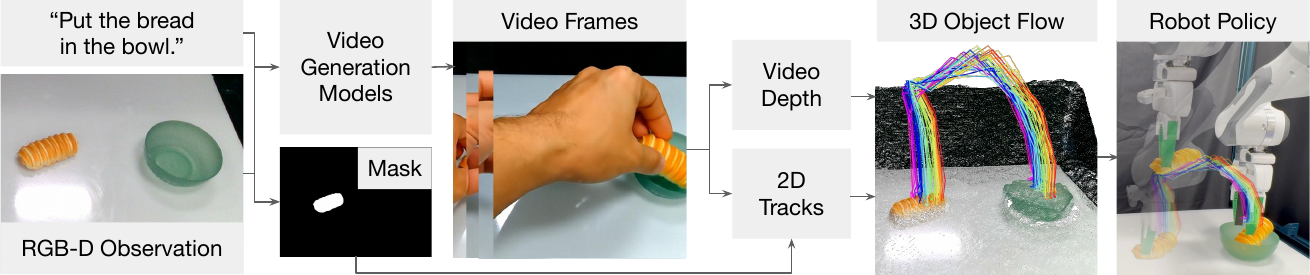}
    \caption{ \textbf{An overview of \algo.} Given a task instruction and an initial RGB-D observation, an image-to-video model synthesizes video frames conditioned on the instruction. We additionally obtain object masks, video depth, and point tracking from vision foundation models, which are used to reconstruct \flow. Finally, a robot policy generates executable actions that track the \flow using trajectory optimization or reinforcement learning.}
    \label{fig:particle_tracking}
    \vspace{-0.5em}
\end{figure*}

Herein, we introduce the problem formulation of \algo in Sec.~\ref{sec:problem}.
Leveraging \flow as an interface, we subsequently discuss how to extract \flow from video generations in Sec.~\ref{sec:particle_tracking} and how to plan actions with \flow for manipulation in Sec.~\ref{sec:planning}.

\subsection{Problem Formulation}
\label{sec:problem}

Given a task instruction $\ell$, an initial RGB-D observation $(I_0 \in \mathbb{R}^{H\!\times\!W\!\times\!3},\, D_0 \in \mathbb{R}^{H\!\times\!W})$, and a known camera projection $\Pi$ (intrinsics and extrinsics to the robot frame), our goal is to output an action sequence $u_{0:H-1} \in \mathcal{U}^H$ that accomplishes the task by following an object motion inferred from a generated video. We make no assumption about a specific action parameterization: $\mathcal{U}$ may represent motion primitives, end-effector poses, or low-level controls.

\textbf{Extracting 3D Object Flow.} From $(I_0,\ell)$, an image-to-video model produces frames $\{V_t\}_{t=1}^T$, and a video-depth estimator provides a per-frame depth sequence $\{Z_t\}_{t=1}^T$. Using a binary mask $M$ of the task-relevant object, and the projection $\Pi$, we lift masked image points with $Z_{1:T}$ to obtain an object-centric 3D trajectory $P_{1:T} \in \mathbb{R}^{T\times n\times 3}$ in the robot frame; we refer to $P_{1:T}$ as the 3D object flow.

\textbf{Action Inference with 3D Object Flow.} We represent state as the task-relevant object and the robot: $x_t = (x_t^{\text{obj}}, r_t)$, where $x_t^{\text{obj}} \in \mathbb{R}^{n\times 3}$ are object points and $r_t$ denotes the robot state. Let $f$ be a dynamics model and $\hat{x}_{t+1} = f(\hat{x}_t, u_t)$ with $\hat{x}_0 = x_0$. At each planning step $t$, we use a time-aligned target $\tilde{P}_t \in \mathbb{R}^{n\times 3}$ derived from the video object flow (e.g., via uniform time-warping or nearest-shape matching). We formulate action inference as an optimization problem:
\begin{equation*}
\label{eq:problem}
\begin{aligned}
\min_{\{u_t \in \mathcal{U}\}} \ \ & \sum_{t=0}^{H-1} \lambda_{\text{task}}\big(\hat{x}^{\text{obj}}_t, \tilde{P}_t\big) + \lambda_{\text{control}}(\hat{x}_t, u_t) \\
\text{s.t.}\ \ & \hat{x}_{t+1} = f(\hat{x}_t, u_t), \quad \hat{x}_0 = x_0,\\
                 & \lambda_{\text{task}}\big(\hat{x}^{\text{obj}}_t, \tilde{P}_t\big) = \sum_{i=1}^n \big\|\hat{x}^{\text{obj}}_t[i]-\tilde{P}_t[i]\big\|_2^2,
\end{aligned}
\end{equation*}
Section~\ref{sec:planning} instantiates $\mathcal{U}$ and $f$ for different domains.

\subsection{Extracting 3D Object Flows from Videos}
\label{sec:particle_tracking}

\textbf{Video Generation:} Given the task language instruction $\ell$ and an RGB image of the workspace without the robot visible $I_0\!\in\!\mathbb{R}^{H \times W \times 3}$, \algo uses an off-the-shelf image-to-video generation model
to produce an RGB video $\{V_t\}_{t=1}^T$ with $V_t\!\in\!\mathbb{R}^{H \times W \times 3}$, showing the task being performed. We do not include the robot in the initial frame or mention the robot in the text prompt, as we empirically find that current image-to-video generation models not specifically finetuned on robotics data tend to produce physically implausible fine-grained interactions, and consequently have worse object trajectories (for details see Appendix~\ref{sec:vid_gen_prompts}).

\textbf{Video Depth Estimation:} Given the generated video, \algo leverages SpatialTrackerV2~\cite{xiao2025spatialtracker,yang2024depth} to estimate per-frame depth $\{\tilde{Z}_t\}_{t=1}^T$, $\tilde{Z}_t\!\in\!\mathbb{R}^{H \times W}$. Due to the scale-shift ambiguity of monocular video, we compute global $(s^\star,b^\star)$ by aligning the first frame to the initial depth $D_0$ from the robot, and obtain calibrated depths $Z_t = s^\star\tilde{Z}_t + b^\star$.

\textbf{3D Object Flow Extraction:} \flow aims to produce 3D trajectories $P_{1:T}\!\in\!\mathbb{R}^{T \times n \times 3}$ with visibilities $V\!\in\!\{0,1\}^{T\times n}$ for the task-relevant object. We first localize the relevant object using Grounding DINO~\cite{liu2023grounding} to produce a bounding box from $(I_0,\ell)$, and then use the box to prompt SAM 2~\cite{ravi2024sam2} for a binary mask. From the masked region at $t{=}1$, we sample $n$ pixels and track them across the video with CoTracker3~\cite{karaev2024cotracker3} to obtain 2D trajectories $c_i^t$ and visibilities $v_i^t$. Visible points are lifted to 3D using the calibrated depths and camera intrinsics/extrinsics, producing $P_{1:T}$ as in Sec.~\ref{sec:problem}.

\subsection{Action Inference with 3D Object Flow}
\label{sec:planning}

\textbf{Simulated Push-T Domain.} For tasks involving non-prehensile manipulation, such as the Push-T task, \algo uses a push skill primitive parameterized with the start push position on a flat table $(c_x, c_y)$, a unit push direction $(\Delta c_x, \Delta c_y)$, and the distance of the push $d$. In this setting, we learn a forward dynamics model that takes as input a feature-augmented particles $\Tilde{x}_t \in \mathbb{R}^{N \times 14}$ of the whole scene and produces $\Delta \hat{x}_{t+1}$, the delta of positions of each point for the next timestep. The features associated with each point consist of the position, RGB color, and normal vector as determined from camera observations along with the push parameters (as further elaborated in Appendix~\ref{sec:particle_dynamics_details}).

To optimize the 3D object flow following cost with the learned dynamics model, we use random-shooting, where $r$ push skill parameters are randomly sampled such that all pushes will make contact with the object of interest at different points and directions. Then, we select the push skill parameter out of $r$ ones that has the least cost according to the predicted point positions from the dynamics model. For determining which timestep from the video should be used in the cost function, we find the timestep $t^{\star}$ where the points of the relevant object in the trajectory are closest to the current observed points with further details in Appendix~\ref{sec:pusht_planning}.

\textbf{Real-World Domain.} We use absolute end-effector poses as the action space and a rigid-grasp dynamics model for the real robot domain. With this combination, we first proceed to grasp the desired part from the relevant object, and then use the dynamics model and point-flow following objective to move the end-effector such that the grasped part moves in a fashion similar to the video. We use AnyGrasp~\cite{fang2023anygrasp} to propose candidate grasps on the object of interest, but these grasps may not lie exactly on the desired part. We select the grasp closest to the thumb as detected from the video by HaMer~\cite{pavlakos2024reconstructing}, as we observe that in generated videos, the hand tends to interact with the relevant part of an object, such as the handle.

The rigid-grasp dynamics model assumes that the grasped part is rigid; consequently, the prediction of point positions to the next timestep can be described by a sequence of rigid transformations for the grasped subset, while non-grasped points remain unchanged. To produce a trajectory of end-effector poses, \algo directly optimizes the objective using PyRoki~\cite{kim2025pyroki}, incorporating pose smoothness and reachability costs as $\lambda_{\text{control}}$ in addition to the 3D object-flow following cost $\lambda_{\text{task}}$ as in Appendix~\ref{sec:real_world_planning}.

\textbf{Simulated Door Opening Domain.}
For the Door Opening task, we use reinforcement learning to learn a sensimotor policy which moves the object according to the \flow with SAC~\cite{haarnojaSAC2018}. This approach can be viewed as using the simulator as a dynamics model for compiling the optimization process prescribed in Eq.~\ref{eq:problem} into a parametric policy in an offline fashion, using \flow as the reward function. Depending on the embodiment used, the action space consists of delta end-effector poses and delta joint angles for the gripper or dexterous hand. The reward function used consists of one term which encourages the end-effector to move toward the current mean object particle position along with another term for encouraging matching the \flow, $\frac{t^{\star}}{t_{\text{end}}}$, where $t^{\star}$ is the timestep with the closest particle positions to the 3D object flow and $t_{\text{end}}$ is the last timestep of the object motion (further details in Appendix~\ref{sec:open_door_rl}).
By training policies with such an object-centric reward, we observe different strategies emerge to accomplish the same object motion across embodiments, including quadruped manipulators, humanoids with dexterous hands, and fixed-base arms with parallel grippers.

\section{Experiments}
We seek to answer the following research questions through our experiments:
$\mathbf{\mathcal{Q}1}$: What are properties of 3D object flow when used as an interface to bridge videos and robot control? 
$\mathbf{\mathcal{Q}2}$: How does \algo perform compared to alternative interfaces?
$\mathbf{\mathcal{Q}3}$: How effective is \flow as a reward for learning sensimotor policies?
$\mathbf{\mathcal{Q}4}$: How does the choice of video model affect \algo in simulation and in real-world tasks?
$\mathbf{\mathcal{Q}5}$: How does the choice of dynamics model affect the performance of \algo?

\subsection{Tasks}
\begin{figure}[!tbp]
    \centering
    \includegraphics[width=0.96\linewidth]{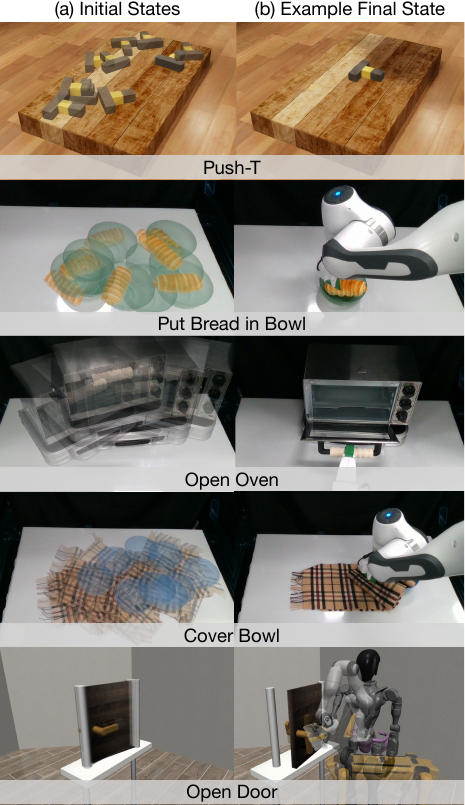}
    \caption{\textbf{Evaluation Tasks.} (a) The initial states of each task used in the evaluation trials. (b) For one of the initial states, the corresponding final state after the robot performs the task. For Push-T only, the desired final state is the same for all initial states.}
    \label{fig:tasks}
\end{figure}

For evaluation, we consider several tasks involving different types of objects and manipulation strategies (Fig.~\ref{fig:tasks}).

\subsubsection{Push-T} 
We develop a simulated Push-T task in OmniGibson~\cite{li2024behavior1k}, where a T-shaped block is placed with a random position and yaw angle on the wooden platform. It is a success if the T-block ends up within 2cm translation and 15 degrees rotation from the goal position, where the T-shaped block is in the center of the board facing forwards.

\subsubsection{Put Bread in Bowl}
The Put Bread in Bowl task consists of a fake piece of bread and a green bowl placed randomly on the workspace. A trial is a success if the bread is inside the bowl at the end.

\subsubsection{Open Oven}
In the Open Oven task, the toaster oven is randomly placed in a semi-circular arc with the orientation towards the base of the robot. A trial is considered a success if the opening angle is at least 60 degrees.

\subsubsection{Cover Bowl}
The Cover Bowl task starts with a folded scarf placed at a random position and orientation on the workspace, with a blue bowl placed next to it. The trial is a success if the scarf is covering at least 25\% of the top of the bowl after the robot finishes execution.

\subsubsection{Open Door}
In the Open Door task from Robosuite~\cite{robosuite2020}, a door is placed at random positions and orientations on top of a table. Rotating the handle and pulling the door open by at least $17^\circ$ without timing out is a success.

\subsection{Properties of 3D Object Flow
as a Video-Control Interface}

\begin{figure}[!tbp]
    \centering
    \includegraphics[width=\linewidth]{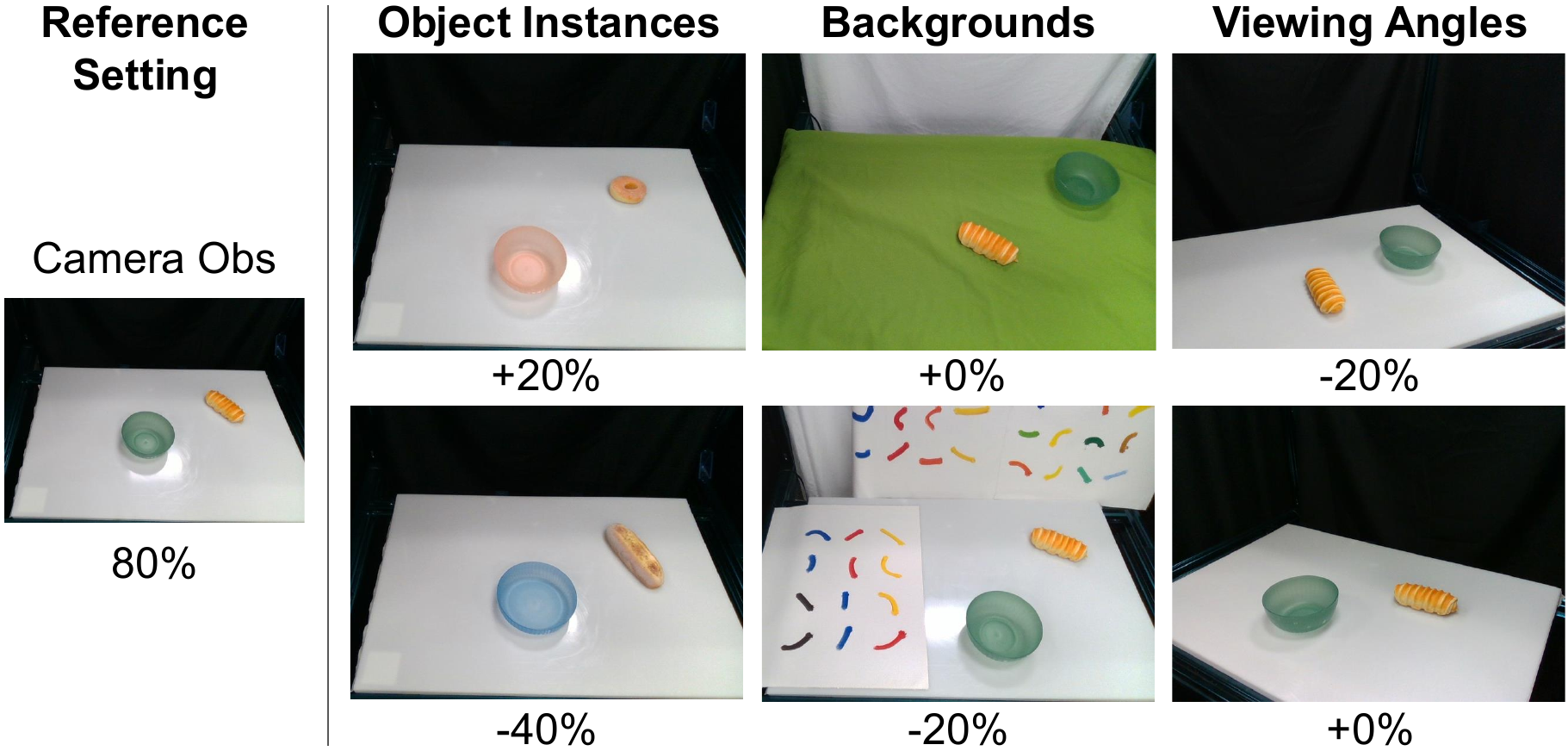}
    \caption{\textbf{Robustness evaluations.} Relative performance across instance, background, and task variations, showing \algo remains robust under various different settings.}
    \label{fig:generalization}
\end{figure}

\begin{figure}[!tbp]
    \centering
    \includegraphics[width=\linewidth]{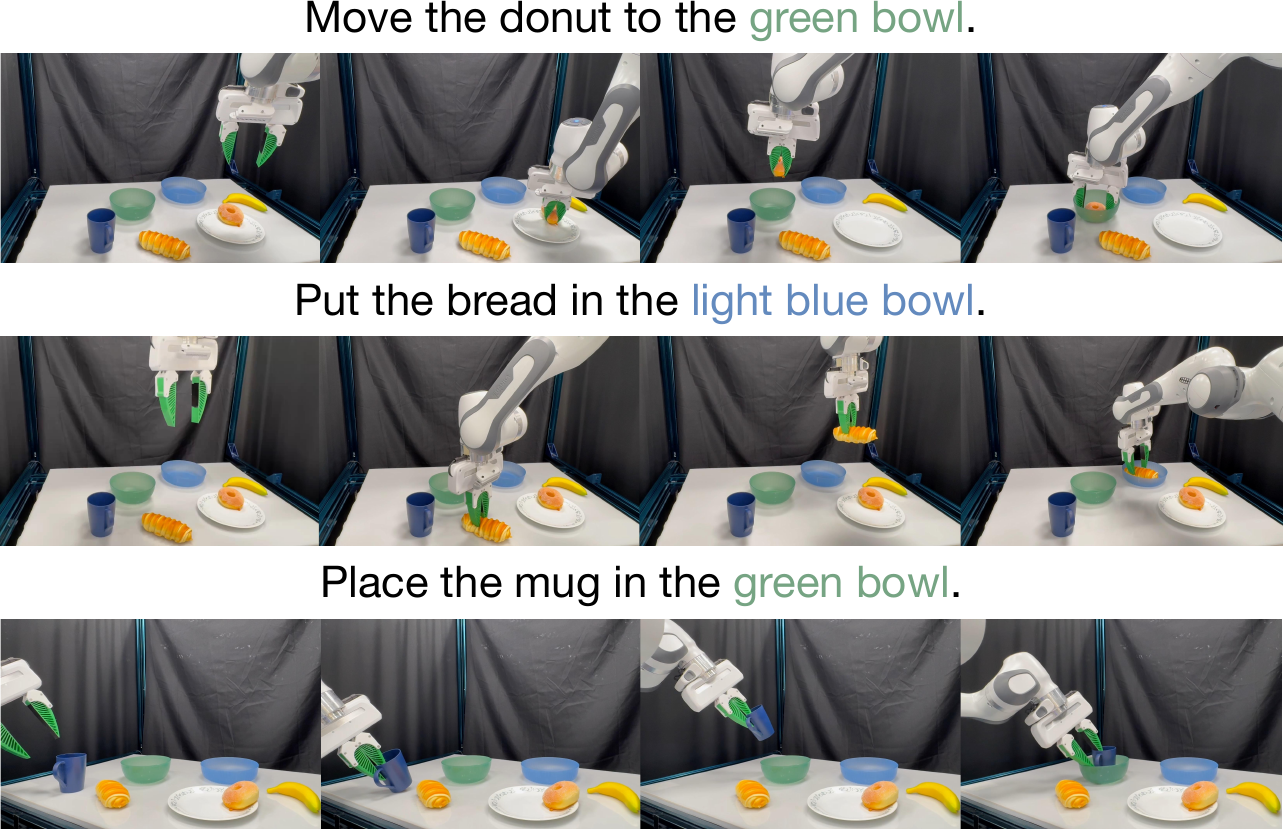}
    \caption{\textbf{Multiple tasks in the same scene.} With different language goals, \algo adapts object-flow targets to produce distinct behaviors in the same environment.}
    \label{fig:same_scene_multi_task}
    \vspace{-0.5em}
\end{figure}

We run \algo on the simulated Push-T task with Wan2.1~\cite{wan2025} as the video generation model. For this task only, we allow prompting with a goal image of the T-block in the center, similar to the example final state in Fig.~\ref{fig:tasks}, as this task requires accurate movement. We consider 10 different initial states with the same final state. We then proceed to run 10 trials for each initial configuration on different seeds to properly evaluate performance due to the stochastic nature of random shooting, yielding 100 total trials. The result appears in Table~\ref{tab:push_t_ablation}. We note that in this task, 6 generated videos included substantial morphing of the T-block, which consequently ruined the tracking and downstream execution. 

We evaluate \algo's ability to perform manipulation with different types of objects with Veo 3~\cite{Veo3TechReport2025} as the video generation model in the real world for 10 trials per task, and report the results in Table~\ref{tab:real_results}. 

To assess \algo's generalization to different object instances, backgrounds, and viewing angles, we conduct an additional five trials each for six different scenarios, as shown in Fig.~\ref{fig:generalization}. Compared to the reference setting, with exception of putting a large piece of bread, there is not a significant drop off in performance, suggesting that \algo inherits some generalization through the use of video generation models. We additionally show that \algo can perform different tasks from the same scene in Fig.~\ref{fig:same_scene_multi_task} due to the ability of video generation models to follow different task instructions given the same input image. We have additional case studies demonstrating that 3D object flow can be used for downstream manipulation in in-the-wild settings for tasks such as pulling a chair, opening a drawer, sweeping pasta, and recycling a can, as shown in Fig.~\ref{fig:pull_fig} and Appendix~\ref{sec:in_the_wild_desc}.

\begin{figure}[!tbp]
    \centering
    \includegraphics[width=0.96\linewidth]{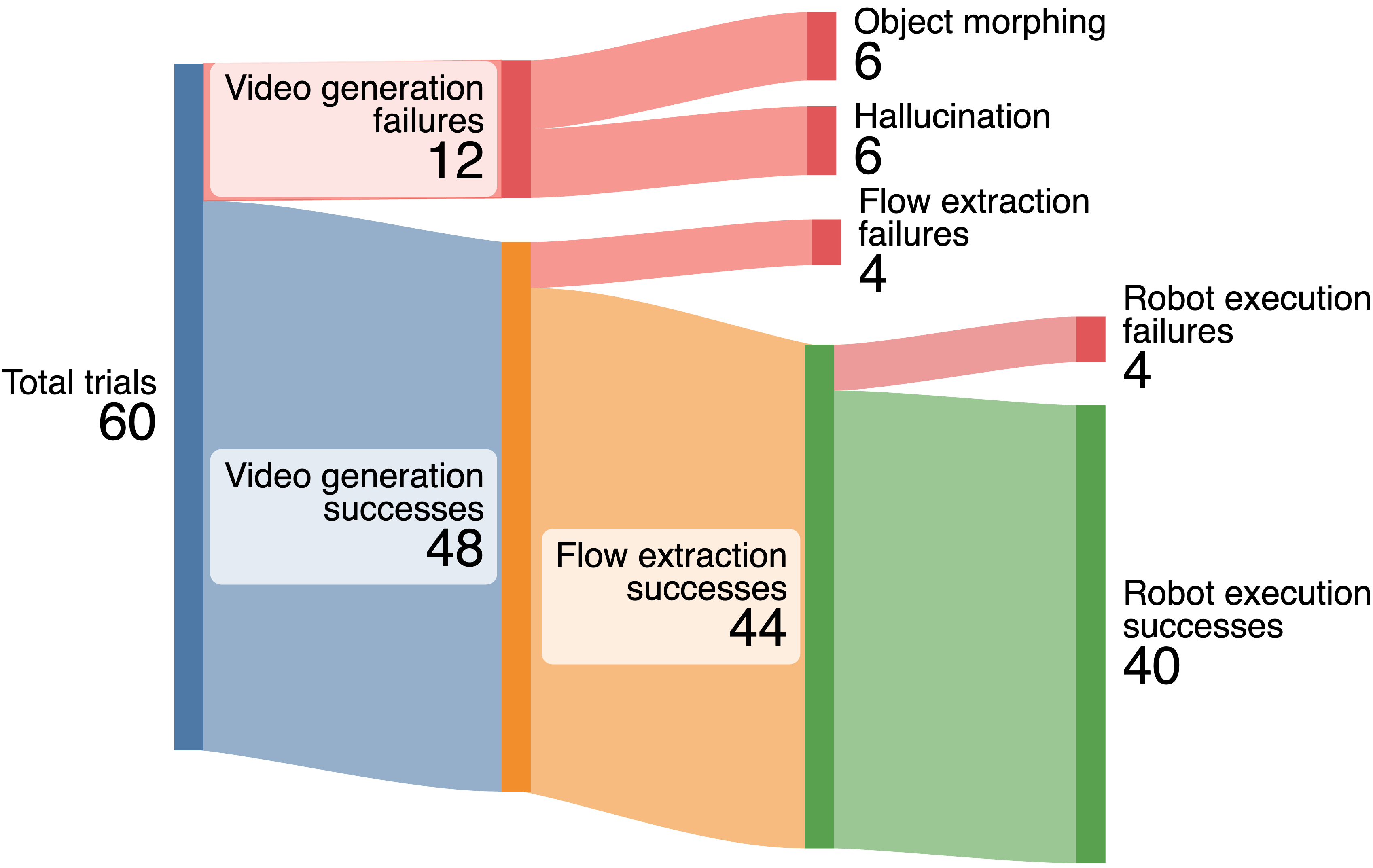}
    \caption{\textbf{Failure breakdown on real-robot experiments.} Common causes include video artifacts (object morphing, object hallucination), tracking errors, and grasp selection mismatches.}
    \label{fig:failures}
    \vspace{-0.3cm}
\end{figure}

For all 60 trials of \algo executed in the real world, we provide a breakdown of failures in Fig.~\ref{fig:failures}. In the 12 video generation failures, for half the time, the generated video either morphs an object in an implausible way or hallucinates new objects, causing tracking to unreasonably fail or making the robot move an object to an incorrect 3D location. The four flow extraction failures occurred because of severe rotations or objects temporarily going out of view of the camera, leading to tracks with no visibility in the end. The four robot execution failures occurred in the Bowl Covering task, where the robot either did not grasp at the correct point or did not move enough.

\subsection{How does Dream2Flow perform compared to alternative
interfaces?}
For the three tasks in the real world, we consider the following alternative interfaces related to extracting object trajectories from videos:

\begin{table}
    \centering
    \begin{tabular}{c|ccc}
    \toprule
    Task & AVDC & RIGVID & \algo \\
    \midrule
    Bread in Bowl & 7/10 & 6/10 & \textbf{8/10}\\
    Open Oven & 0/10 & 6/10 & \textbf{8/10} \\
    Cover Bowl & 2/10 & 1/10 & \textbf{3/10} \\
    \bottomrule
    \end{tabular}
\caption{\textbf{Comparisons of intermediate representations on real robot.} Dream2Flow outperforms AVDC and RIGVID across three tasks by following 3D object flow rather than rigid transforms alone.}
\label{tab:real_results}
\vspace{-0.3cm}
\end{table}

\subsubsection{AVDC} AVDC~\cite{ko2023learning} leverages generated videos by computing dense optical flow between frames to track points on a rigid object. Then, using the initial depth and the point correspondences, it solves for a sequence of rigid transforms of the relevant object, allowing for trajectory playback after a grasp has occurred. In our implementation, we utilize the video depth corresponding to the tracks from the optical flow, and optimize for rigid transforms relative to the initial frame, as we empirically found that to be less noisy than the original optimization procedure. 

\subsubsection{RIGVID} 
RIGVID~\cite{patel2025rigvid} uses 6D object pose tracking to generate a rigid object trajectory from a generated video. Since it is ill-defined to have such a pose for deformable objects, we do not use a 6D pose tracker, but instead adapt their approach by solving for a rigid pose transformation between the initial 3D points and visible 3D points from our 3D Object Flow computation in the same manner as AVDC. 

We present the results in Table~\ref{tab:real_results}. While AVDC does a reasonable job at tracking the bread, the dense optical flow does not keep up with the motion of the oven, resulting in insufficient motion. Under certain circumstances, there are only a few visible points for RIGVID and AVDC, making transform estimation noisy. Attempting to follow the noisy transform trajectories leads to execution failures or optimization instability. \algo is less affected by this issue, because there is typically not a heavy cost when most of the points are occluded, allowing the planned end-effector poses to smoothly move between areas of high point visibility. The Cover Bowl task remains a challenge for AVDC and RIGVID, as in addition to video generation and tracking failures, the transform estimates are incorrect since the points are now on a deformable object.

\subsection{How effective is 3D
object flow as a reward for learning sensimotor policies?}
\begin{figure}[!tbp]
    \centering
    \includegraphics[width=0.96\linewidth]{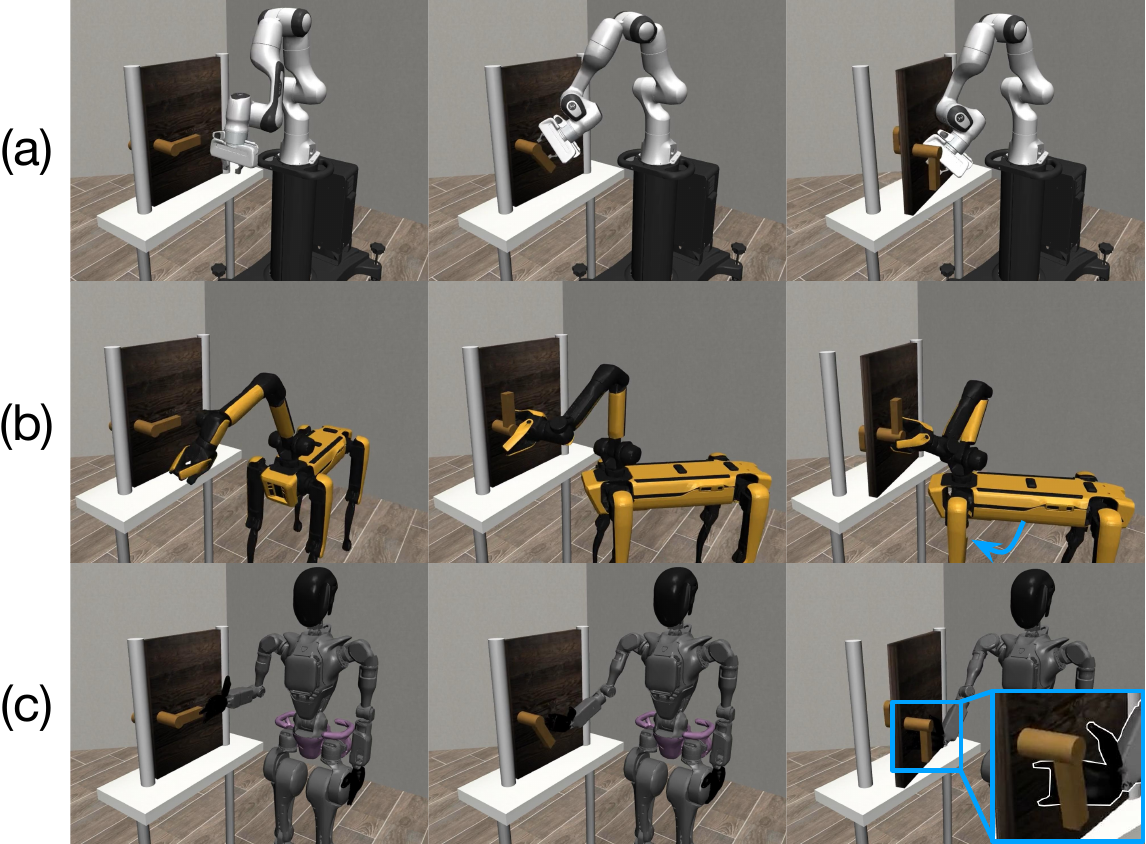}
    \caption{\textbf{Rollouts from policies trained using 3D object flow as a reward.} Different embodiments such as the (a) Panda, (b) Spot, or (c) GR1 use different strategies to open the door. The Spot is able move its base for better reachability while the GR1 uses the area between its fingers and palm to pull for better stability.}
    \label{fig:rl_rollout_vis}
    \vspace{-0.5em}
\end{figure}

The 3D object flow extracted by \algo can be used in an RL reward for training policies across different embodiments. We evaluate SAC~\cite{haarnojaSAC2018} policies trained using handcrafted object state reward from Robosuite and 3D object flow rewards on a Franka Panda, a Spot with floating base, and a GR1 with the right arm only across 100 random door positions, and report the results in Table~\ref{tab:rl_results} with corresponding episode visualizations in Figure~\ref{fig:rl_rollout_vis}. The policies trained with the object state and 3D object flow reward have comparable performance across all the embodiments. The learned strategies are different between the embodiments, as the Spot is able to move its base for better reachability and kinematic range, and the GR1 uses the area between the fingers and palm for pulling the door as opposed to individual fingers for better stability.

\begin{table}[!tbp]
    \centering
    \begin{tabular}{c|ccc}
    \toprule
    Reward Type & Franka & Spot & GR1 \\
    \midrule
    Object State & 99/100 & 99/100 & \textbf{96/100}\\
    3D Object Flow & \textbf{100/100} & \textbf{100/100} & 94/100 \\
    \bottomrule
    \end{tabular}
\caption{\textbf{Comparison of policies trained using different rewards.} The policies trained using the 3D object flow reward perform comparably to those trained with the object state reward across different embodiments.}
\label{tab:rl_results}
\end{table}

\subsection{How does the choice of video model affect Dream2Flow in simulation and in real-world tasks?}
\begin{table}[!thbp]
    \centering
    \begin{tabular}{c|cc}
    \toprule
    Video Generation Model & Push-T & Open Oven\\
    \midrule
    Wan2.1~\cite{wan2025} & \textbf{52/100} & 2/10\\
    Kling 2.1 & 31/100 & 4/10\\
    Veo 3~\cite{Veo3TechReport2025} &  - & \textbf{8/10} \\
    \bottomrule
    \end{tabular}
\caption{\textbf{Effect of video generator.} Veo 3 excels on real-world domains such as ``Open Oven'', while Wan 2.1 performs better on simulated domains such as ``Push-T''.}
\label{tab:vid_gen_comparison}
\vspace{-0.3cm}
\end{table}

To evaluate how well different video generation models capture physically plausible object trajectories, we run \algo on the simulated Push-T task and the real Open Oven task using three models: Wan2.1~\cite{wan2025}, Kling 2.1, and Veo 3~\cite{Veo3TechReport2025}. The results are shown in Table~\ref{tab:vid_gen_comparison}. Note that there are no results for Veo 3 on Push-T because at the time of evaluation, Veo 3 did not support prompting with a goal image.

For the Push-T task, Kling 2.1 had more videos that had substantial morphing, throwing off the tracking, resulting in more downstream failures. For the Open Oven task, Wan2.1 tends to produce more videos with substantial camera motion, which violates the still camera assumption. Additionally, both Kling 2.1 and Wan2.1 produce videos where the direction of articulation is incorrect, such as revolving around the wrong axis, leading to far more failures than Veo 3.

\subsection{How does the choice of dynamics model affect the performance of Dream2Flow?}
\begin{table}
    \centering
    \begin{tabular}{ccc}
    \toprule
    Dynamics Model Type & Success Rate\\
    \midrule
    Pose &  12/100\\
    Heuristic & 17/100\\
    Particle & \textbf{52/100}\\
    \bottomrule
    \end{tabular}
\caption{\textbf{Dynamics model ablation.} Particle dynamics substantially outperform pose and heuristic models, highlighting the importance of per-point predictions.}
\label{tab:push_t_ablation}
\end{table}

For the Push-T task, we compare the effect of different dynamics models used for planning. In addition to the particle-based dynamics model, we consider another learned dynamics model which takes in the block pose and push skill parameters and predicts the delta pose of the T-block, trained on the same data as the particle based dynamics model, as well as a heuristic dynamics model, which translates the points of the T-block in the direction and amount of the push without any rotation. We present the results in Table~\ref{tab:push_t_ablation}, and find that the particle representation for this dynamics model is crucial to ensure success, as despite having the same 3D object flow guidance, the pose and heuristic based dynamics models could not sufficiently account for the rotation needed.  

\section{Conclusion}
We presented \algo, a simple, general interface that turns text-conditioned video predictions into executable robotic actions by reconstructing and tracking 3D object flow. This decouples what should happen in the world (task-relevant object motion and state change) from how a particular embodiment realizes it under kinematic, dynamic, and morphology constraints. Built entirely from off-the-shelf video generation and perception tools, \algo solves open-world manipulation tasks in simulation and on real robots across rigid, articulated, deformable, and granular objects using only RGB-D observations and language, without task-specific demonstrations. Experiments show consistent gains over trajectory baselines derived from dense optical flow or rigid pose transforms, robustness to variations in instances, backgrounds, and viewpoints, and the importance of both the upstream video model and downstream dynamics choice (with particle dynamics proving most reliable). A failure analysis highlights current bottlenecks, such as video artifacts (morphing, hallucinations), occlusion-induced tracking dropouts, and grasp selection mismatches as concrete directions for improvement.  Overall, our results indicate that 3D object flow is a scalable bridge from open-ended video generation to robot control in unstructured environments.

\section*{Acknowledgments}
This work is in part supported by the Stanford Institute for Human-Centered AI (HAI), the Schmidt Futures Senior Fellows grant, ONR MURI N00014-21-1-2801, ONR MURI N00014-22-1-2740.

\bibliographystyle{IEEEtran}
\bibliography{ref}

@inproceedings{xiao2025spatialtracker,
  title={SpatialTrackerV2: 3D Point Tracking Made Easy},
  author={Xiao, Yuxi and Wang, Jianyuan and Xue, Nan and Karaev, Nikita and Makarov, Iurii and Kang, Bingyi and Zhu, Xin and Bao, Hujun and Shen, Yujun and Zhou, Xiaowei},
  booktitle={ICCV},
  year={2025}
}

@article{weng2021fabricflownet,
  title={FabricFlowNet: Bimanual Cloth Manipulation with a Flow-based Policy},
  author={Weng, Thomas and Bajracharya, Sujay and Wang, Yufei and Agrawal, Khush and Held, David},
  journal={arXiv preprint arXiv:2111.05623},
  year={2021}
}

@article{goyal2022ifor,
  title={IFOR: Iterative Flow Minimization for Robotic Object Rearrangement},
  author={Goyal, Ankit and Mousavian, Arsalan and Paxton, Chris and Chao, Yu-Wei and Okorn, Brian and Deng, Jia and Fox, Dieter},
  journal={arXiv preprint arXiv:2202.00732},
  year={2022}
}

@article{seita2023toolflownet,
  title={ToolFlowNet: Robotic Manipulation with Tools via Predicting Tool Flow from Point Clouds},
  author={Seita, Daniel and Wang, Yufei and Shetty, Sarthak J and Li, Edward Yao and Erickson, Zackory and Held, David},
  journal={arXiv preprint arXiv:2211.09006},
  year={2022}
}

@article{chen2024g3flow,
  title={G3Flow: Generative 3D Semantic Flow for Pose-aware and Generalizable Object Manipulation},
  author={Chen, Tianxing and Mu, Yao and Liang, Zhixuan and Chen, Zanxin and Peng, Shijia and Chen, Qiangyu and Xu, Mingkun and Hu, Ruizhen and Zhang, Hongyuan and Li, Xuelong and Luo, Ping},
  journal={arXiv preprint arXiv:2411.18369},
  year={2024}
}

@inproceedings{eisner2022flowbot3d,
  title={FlowBot3D: Learning 3D Articulation Flow to Manipulate Articulated Objects},
  author={Eisner, Ben and Zhang, Harry and Held, David},
  booktitle={Robotics: Science and Systems (RSS)},
  year={2022}
}

@article{guo2025flowdreamer,
  title={FlowDreamer: A RGB-D World Model with Flow-based Motion Representations for Robot Manipulation},
  author={Guo, Jun and Ma, Xiaojian and Wang, Yikai and Yang, Min and Liu, Huaping and Li, Qing},
  journal={arXiv preprint arXiv:2505.10075},
  year={2025}
}

@misc{nvidia2025zeromsf,
  title={ZeroMSF: Zero-shot Monocular Scene Flow in the Wild},
  howpublished={NVIDIA Research Project Page},
  year={2025},
}

@article{sander2025tofsceneflow,
  title={Estimating Scene Flow in Robot Surroundings with Distributed Miniaturized Time-of-Flight Sensors},
  author={Sander, Jack and Caroleo, Giammarco and Albini, Alessandro and Maiolino, Perla},
  journal={arXiv preprint arXiv:2504.02439},
  year={2025}
}

@article{liu2018flownet3d,
  title={FlowNet3D: Learning Scene Flow in 3D Point Clouds},
  author={Liu, Xingyu and Qi, Charles R and Guibas, Leonidas J},
  journal={arXiv preprint arXiv:1806.01411},
  year={2018}
}

@article{byravan2016se3nets,
  title={SE3-Nets: Learning Rigid Body Motion using Deep Neural Networks},
  author={Byravan, Arunkumar and Fox, Dieter},
  journal={arXiv preprint arXiv:1606.02378},
  year={2016}
}

@article{shao2018motionseg,
  title={Motion-based Object Segmentation based on Dense RGB-D Scene Flow},
  author={Shao, Lin and Shah, Parth and Dwaracherla, Vikranth and Bohg, Jeannette},
  journal={arXiv preprint arXiv:1804.05195},
  year={2018}
}

@article{tang2023rftrans,
  title={RFTrans: Leveraging Refractive Flow of Transparent Objects for Surface Normal Estimation and Manipulation},
  author={Tang, Tutian and Liu, Jiyu and Zhang, Jieyi and Fu, Haoyuan and Xu, Wenqiang and Lu, Cewu},
  journal={arXiv preprint arXiv:2311.12398},
  year={2023}
}

@article{shorinwa2024splatmover,
  title={Splat-MOVER: Multi-Stage, Open-Vocabulary Robotic Manipulation via Editable Gaussian Splatting},
  author={Shorinwa, Ola and Tucker, Johnathan and Smith, Aliyah and Swann, Aiden and Chen, Timothy and Firoozi, Roya and Kennedy III, Monroe and Schwager, Mac},
  journal={arXiv preprint arXiv:2405.04378},
  year={2024}
}

@article{jiang2024roboexp,
  title={RoboEXP: Action-Conditioned Scene Graph via Interactive Exploration for Robotic Manipulation},
  author={Jiang, Hanxiao and Huang, Binghao and Wu, Ruihai and Li, Zhuoran and Garg, Shubham and Nayyeri, Hooshang and Wang, Shenlong and Li, Yunzhu},
  journal={arXiv preprint arXiv:2402.15487},
  year={2024}
}

@article{kerr2024rsrd,
  title={Robot See Robot Do: Imitating Articulated Object Manipulation with Monocular 4D Reconstruction},
  author={Kerr, Justin and Kim, Chung Min and Wu, Mingxuan and Yi, Brent and Wang, Qianqian and Goldberg, Ken and Kanazawa, Angjoo},
  journal={arXiv preprint arXiv:2409.18121},
  year={2024}
}

@article{duisterhof2024deformgs,
  title={DeformGS: Scene Flow in Highly Deformable Scenes for Deformable Object Manipulation},
  author={Duisterhof, Bardienus P. and Mandi, Zhao and Yao, Yunchao and Liu, Jia-Wei and Seidenschwarz, Jenny and Shou, Mike Zheng and Ramanan, Deva and Song, Shuran and Birchfield, Stan and Wen, Bowen and Ichnowski, Jeffrey},
  journal={arXiv preprint arXiv:2312.00583},
  year={2024}
}

@article{guo2025actionsink,
  title={ActionSink: Toward Precise Robot Manipulation with Dynamic Integration of Action Flow},
  author={Guo, Shanshan and Liang, Xiwen and Lin, Junfan and Zhuang, Yuzheng and Lin, Liang and Liang, Xiaodan},
  journal={arXiv preprint arXiv:2508.03218},
  year={2025}
}

@article{bharadhwaj2024gen2act,
  title={Gen2Act: Human Video Generation in Novel Scenarios enables Generalizable Robot Manipulation},
  author={Bharadhwaj, Homanga and Dwibedi, Debidatta and Gupta, Abhinav and Tulsiani, Shubham and Doersch, Carl and Xiao, Ted and Shah, Dhruv and Xia, Fei and others},
  journal={arXiv preprint arXiv:2409.16283},
  year={2024}
}

@article{patel2025rigvid,
  title={Robotic Manipulation by Imitating Generated Videos Without Physical Demonstrations},
  author={Patel, Shivansh and Mohan, Shraddhaa and Mai, Hanlin and Jain, Unnat and Lazebnik, Svetlana and Li, Yunzhu},
  journal={arXiv preprint arXiv:2507.00990},
  year={2025}
}

@article{he2025manitrend,
  title={ManiTrend: Bridging Future Generation and Action Prediction with 3D Flow for Robotic Manipulation},
  author={He, Yuxin and Nie, Qiang},
  journal={arXiv preprint arXiv:2502.10028},
  year={2025}
}

@article{chen2025ecflow,
  title={EC-Flow: Enabling Versatile Robotic Manipulation from Action-Unlabeled Videos via Embodiment-Centric Flow},
  author={Chen, Yixiang and Li, Peiyan and Huang, Yan and Yang, Jiabing and Chen, Kehan and Wang, Liang},
  journal={arXiv preprint arXiv:2507.06224},
  year={2025}
}

@article{zhi2025flowaction,
  title={3DFlowAction: Learning Cross-Embodiment Manipulation from 3D Flow World Model},
  author={Zhi, Hongyan and Chen, Peihao and Zhou, Siyuan and Dong, Yubo and Wu, Quanxi and Han, Lei and Tan, Mingkui},
  journal={arXiv preprint arXiv:2506.06199},
  year={2025}
}

@article{karaev2024cotracker3,
    author    = {Nikita Karaev and Iurii Makarov and Jianyuan Wang and Natalia Neverova and Andrea Vedaldi and Christian Rupprecht},
    title     = {{CoTracker3}: Simpler and Better Point Tracking by Pseudo-Labelling Real Videos},
    journal   = {arxiv},
    year      = {2024}
}

@article{fang2023anygrasp,
  title={AnyGrasp: Robust and Efficient Grasp Perception in Spatial and Temporal Domains},
  author = {Fang, Hao-Shu and Wang, Chenxi and Fang, Hongjie and Gou, Minghao and Liu, Jirong and Yan, Hengxu and Liu, Wenhai and Xie, Yichen and Lu, Cewu},
  journal={IEEE Transactions on Robotics (T-RO)},
  year={2023}
}

@article{wan2025,
      title={Wan: Open and Advanced Large-Scale Video Generative Models}, 
      author={Team Wan and Ang Wang and Baole Ai and Bin Wen and Chaojie Mao and et al.},
      journal = {arXiv preprint arXiv:2503.20314},
      year={2025}
}

@misc{Veo3TechReport2025,
  title        = {{Veo-3 Technical Report}},
  author       = {{Google DeepMind}},
  year         = {2025},
  type         = {Technical Report},
  url          = {https://storage.googleapis.com/deepmind-media/veo/Veo-3-Tech-Report.pdf},
}

@inproceedings{pavlakos2024reconstructing,
  title={Reconstructing Hands in 3{D} with Transformers},
  author={Pavlakos, Georgios and Shan, Dandan and Radosavovic, Ilija and Kanazawa, Angjoo and Fouhey, David and Malik, Jitendra},
  booktitle={CVPR},
  year={2024}
}

@article{liu2023grounding,
  title={Grounding dino: Marrying dino with grounded pre-training for open-set object detection},
  author={Liu, Shilong and Zeng, Zhaoyang and Ren, Tianhe and Li, Feng and Zhang, Hao and Yang, Jie and Li, Chunyuan and Yang, Jianwei and Su, Hang and Zhu, Jun and others},
  journal={arXiv preprint arXiv:2303.05499},
  year={2023}
}

@article{ravi2024sam2,
  title={SAM 2: Segment Anything in Images and Videos},
  author={Ravi, Nikhila and Gabeur, Valentin and Hu, Yuan-Ting and Hu, Ronghang and Ryali, Chaitanya and Ma, Tengyu and Khedr, Haitham and R{\"a}dle, Roman and Rolland, Chloe and Gustafson, Laura and Mintun, Eric and Pan, Junting and Alwala, Kalyan Vasudev and Carion, Nicolas and Wu, Chao-Yuan and Girshick, Ross and Doll{\'a}r, Piotr and Feichtenhofer, Christoph},
  journal={arXiv preprint arXiv:2408.00714},
  url={https://arxiv.org/abs/2408.00714},
  year={2024}
}

@article{garrett2021integrated,
  title={Integrated task and motion planning},
  author={Garrett, Caelan Reed and Chitnis, Rohan and Holladay, Rachel and Kim, Beomjoon and Silver, Tom and Kaelbling, Leslie Pack and Lozano-P{\'e}rez, Tom{\'a}s},
  journal={Annual review of control, robotics, and autonomous systems},
  volume={4},
  number={1},
  pages={265--293},
  year={2021},
  publisher={Annual Reviews}
}

@article{zhao2024survey,
  title={A survey of optimization-based task and motion planning: From classical to learning approaches},
  author={Zhao, Zhigen and Cheng, Shuo and Ding, Yan and Zhou, Ziyi and Zhang, Shiqi and Xu, Danfei and Zhao, Ye},
  journal={IEEE/ASME Transactions on Mechatronics},
  year={2024},
  publisher={IEEE}
}

@inproceedings{toussaint2015logic,
  title={Logic-Geometric Programming: An Optimization-Based Approach to Combined Task and Motion Planning.},
  author={Toussaint, Marc},
  booktitle={IJCAI},
  pages={1930--1936},
  year={2015}
}

@inproceedings{shridhar2022cliport,
  title={Cliport: What and where pathways for robotic manipulation},
  author={Shridhar, Mohit and Manuelli, Lucas and Fox, Dieter},
  booktitle={Conference on robot learning},
  pages={894--906},
  year={2022},
  organization={PMLR}
}

@inproceedings{shridhar2023perceiver,
  title={Perceiver-actor: A multi-task transformer for robotic manipulation},
  author={Shridhar, Mohit and Manuelli, Lucas and Fox, Dieter},
  booktitle={Conference on Robot Learning},
  pages={785--799},
  year={2023},
  organization={PMLR}
}

@article{black2024pi_0,
  title={pi0: A Vision-Language-Action Flow Model for General Robot Control},
  author={Black, Kevin and Brown, Noah and Driess, Danny and Esmail, Adnan and Equi, Michael and Finn, Chelsea and Fusai, Niccolo and Groom, Lachy and Hausman, Karol and Ichter, Brian and others},
  journal={arXiv preprint arXiv:2410.24164},
  year={2024}
}

@inproceedings{kim2025openvla,
  title={OpenVLA: An Open-Source Vision-Language-Action Model},
  author={Kim, Moo Jin and Pertsch, Karl and Karamcheti, Siddharth and Xiao, Ted and Balakrishna, Ashwin and Nair, Suraj and Rafailov, Rafael and Foster, Ethan P and Sanketi, Pannag R and Vuong, Quan and others},
  booktitle={Conference on Robot Learning},
  pages={2679--2713},
  year={2025},
  organization={PMLR}
}

@inproceedings{zitkovich2023rt,
  title={Rt-2: Vision-language-action models transfer web knowledge to robotic control},
  author={Zitkovich, Brianna and Yu, Tianhe and Xu, Sichun and Xu, Peng and Xiao, Ted and Xia, Fei and Wu, Jialin and Wohlhart, Paul and Welker, Stefan and Wahid, Ayzaan and others},
  booktitle={Conference on Robot Learning},
  pages={2165--2183},
  year={2023},
  organization={PMLR}
}

@inproceedings{lynch2020learning,
  title={Learning latent plans from play},
  author={Lynch, Corey and Khansari, Mohi and Xiao, Ted and Kumar, Vikash and Tompson, Jonathan and Levine, Sergey and Sermanet, Pierre},
  booktitle={Conference on robot learning},
  pages={1113--1132},
  year={2020},
  organization={Pmlr}
}

@article{mendonca2021discovering,
  title={Discovering and achieving goals via world models},
  author={Mendonca, Russell and Rybkin, Oleh and Daniilidis, Kostas and Hafner, Danijar and Pathak, Deepak},
  journal={Advances in Neural Information Processing Systems},
  volume={34},
  pages={24379--24391},
  year={2021}
}

@inproceedings{simeonov2022neural,
  title={Neural descriptor fields: Se (3)-equivariant object representations for manipulation},
  author={Simeonov, Anthony and Du, Yilun and Tagliasacchi, Andrea and Tenenbaum, Joshua B and Rodriguez, Alberto and Agrawal, Pulkit and Sitzmann, Vincent},
  booktitle={2022 International Conference on Robotics and Automation (ICRA)},
  pages={6394--6400},
  year={2022},
  organization={IEEE}
}

@inproceedings{xu2025flow,
  title={Flow as the Cross-domain Manipulation Interface},
  author={Xu, Mengda and Xu, Zhenjia and Xu, Yinghao and Chi, Cheng and Wetzstein, Gordon and Veloso, Manuela and Song, Shuran},
  booktitle={Conference on Robot Learning},
  pages={2475--2499},
  year={2025},
  organization={PMLR}
}

@article{ebert2018visual,
  title={Visual foresight: Model-based deep reinforcement learning for vision-based robotic control},
  author={Ebert, Frederik and Finn, Chelsea and Dasari, Sudeep and Xie, Annie and Lee, Alex and Levine, Sergey},
  journal={arXiv preprint arXiv:1812.00568},
  year={2018}
}

@article{black2023zero,
  title={Zero-shot robotic manipulation with pretrained image-editing diffusion models},
  author={Black, Kevin and Nakamoto, Mitsuhiko and Atreya, Pranav and Walke, Homer and Finn, Chelsea and Kumar, Aviral and Levine, Sergey},
  journal={arXiv preprint arXiv:2310.10639},
  year={2023}
}

@inproceedings{xie2018few,
  title={Few-shot goal inference for visuomotor learning and planning},
  author={Xie, Annie and Singh, Avi and Levine, Sergey and Finn, Chelsea},
  booktitle={Conference on Robot Learning},
  pages={40--52},
  year={2018},
  organization={PMLR}
}

@article{sharma2019third,
  title={Third-person visual imitation learning via decoupled hierarchical controller},
  author={Sharma, Pratyusha and Pathak, Deepak and Gupta, Abhinav},
  journal={Advances in Neural Information Processing Systems},
  volume={32},
  year={2019}
}

@article{huang2023voxposer,
  title={VoxPoser: Composable 3D Value Maps for Robotic Manipulation with Language Models},
  author={Huang, Wenlong and Wang, Chen and Zhang, Ruohan and Li, Yunzhu and Wu, Jiajun and Fei-Fei, Li},
  journal={Proceedings of Machine Learning Research},
  volume={229},
  year={2023},
  publisher={ML Research Press}
}

@inproceedings{huang2025rekep,
  title={ReKep: Spatio-Temporal Reasoning of Relational Keypoint Constraints for Robotic Manipulation},
  author={Huang, Wenlong and Wang, Chen and Li, Yunzhu and Zhang, Ruohan and Fei-Fei, Li},
  booktitle={Conference on Robot Learning},
  pages={4573--4602},
  year={2025},
  organization={PMLR}
}

@article{tang2025uad,
  title={UAD: Unsupervised Affordance Distillation for Generalization in Robotic Manipulation},
  author={Tang, Yihe and Huang, Wenlong and Wang, Yingke and Li, Chengshu and Yuan, Roy and Zhang, Ruohan and Wu, Jiajun and Fei-Fei, Li},
  journal={arXiv preprint arXiv:2506.09284},
  year={2025}
}

@inproceedings{liang2023code,
  title={Code as Policies: Language Model Programs for Embodied Control},
  author={Liang, Jacky and Huang, Wenlong and Xia, Fei and Xu, Peng and Hausman, Karol and Ichter, Brian and Florence, Pete and Zeng, Andy},
  booktitle={2023 IEEE International Conference on Robotics and Automation (ICRA)},
  pages={9493--9500},
  year={2023},
  organization={IEEE}
}

@inproceedings{guzey2025bridging,
  title={Bridging the human to robot dexterity gap through object-oriented rewards},
  author={Guzey, Irmak and Dai, Yinlong and Savva, Georgy and Bhirangi, Raunaq and Pinto, Lerrel},
  booktitle={2025 IEEE International Conference on Robotics and Automation (ICRA)},
  pages={3344--3351},
  year={2025},
  organization={IEEE}
}

@inproceedings{bharadhwaj2024track2act,
  title={Track2act: Predicting point tracks from internet videos enables generalizable robot manipulation},
  author={Bharadhwaj, Homanga and Mottaghi, Roozbeh and Gupta, Abhinav and Tulsiani, Shubham},
  booktitle={European Conference on Computer Vision},
  pages={306--324},
  year={2024},
  organization={Springer}
}

@article{adeniji2025feel,
  title={Feel the Force: Contact-Driven Learning from Humans},
  author={Adeniji, Ademi and Chen, Zhuoran and Liu, Vincent and Pattabiraman, Venkatesh and Bhirangi, Raunaq and Haldar, Siddhant and Abbeel, Pieter and Pinto, Lerrel},
  journal={arXiv preprint arXiv:2506.01944},
  year={2025}
}

@article{haldar2025point,
  title={Point policy: Unifying observations and actions with key points for robot manipulation},
  author={Haldar, Siddhant and Pinto, Lerrel},
  journal={arXiv preprint arXiv:2502.20391},
  year={2025}
}

@article{yin2025object,
  title={Object-centric 3D Motion Field for Robot Learning from Human Videos},
  author={Yin, Zhao-Heng and Yang, Sherry and Abbeel, Pieter},
  journal={arXiv preprint arXiv:2506.04227},
  year={2025}
}

@inproceedings{yuan2025general,
  title={General Flow as Foundation Affordance for Scalable Robot Learning},
  author={Yuan, Chengbo and Wen, Chuan and Zhang, Tong and Gao, Yang},
  booktitle={Conference on Robot Learning},
  pages={1541--1566},
  year={2025},
  organization={PMLR}
}

@article{ko2023learning,
  title={Learning to act from actionless videos through dense correspondences},
  author={Ko, Po-Chen and Mao, Jiayuan and Du, Yilun and Sun, Shao-Hua and Tenenbaum, Joshua B},
  journal={arXiv preprint arXiv:2310.08576},
  year={2023}
}

@article{yu2025genflowrl,
  title={GenFlowRL: Shaping Rewards with Generative Object-Centric Flow in Visual Reinforcement Learning},
  author={Yu, Kelin and Zhang, Sheng and Soora, Harshit and Huang, Furong and Huang, Heng and Tokekar, Pratap and Gao, Ruohan},
  journal={arXiv preprint arXiv:2508.11049},
  year={2025}
}

@article{gao2024flip,
  title={Flip: Flow-centric generative planning as general-purpose manipulation world model},
  author={Gao, Chongkai and Zhang, Haozhuo and Xu, Zhixuan and Cai, Zhehao and Shao, Lin},
  journal={arXiv preprint arXiv:2412.08261},
  year={2024}
}

@article{luo2024grounding,
  title={Grounding video models to actions through goal conditioned exploration},
  author={Luo, Yunhao and Du, Yilun},
  journal={arXiv preprint arXiv:2411.07223},
  year={2024}
}

@article{yang2024video,
  title={Video as the new language for real-world decision making},
  author={Yang, Sherry and Walker, Jacob and Parker-Holder, Jack and Du, Yilun and Bruce, Jake and Barreto, Andre and Abbeel, Pieter and Schuurmans, Dale},
  journal={arXiv preprint arXiv:2402.17139},
  year={2024}
}

@article{mccarthy2025towards,
  title={Towards generalist robot learning from internet video: A survey},
  author={McCarthy, Robert and Tan, Daniel CH and Schmidt, Dominik and Acero, Fernando and Herr, Nathan and Du, Yilun and Thuruthel, Thomas G and Li, Zhibin},
  journal={Journal of Artificial Intelligence Research},
  volume={83},
  year={2025}
}

@article{escontrela2023video,
  title={Video prediction models as rewards for reinforcement learning},
  author={Escontrela, Alejandro and Adeniji, Ademi and Yan, Wilson and Jain, Ajay and Peng, Xue Bin and Goldberg, Ken and Lee, Youngwoon and Hafner, Danijar and Abbeel, Pieter},
  journal={Advances in Neural Information Processing Systems},
  volume={36},
  pages={68760--68783},
  year={2023}
}

@inproceedings{huang2024diffusion,
  title={Diffusion reward: Learning rewards via conditional video diffusion},
  author={Huang, Tao and Jiang, Guangqi and Ze, Yanjie and Xu, Huazhe},
  booktitle={European Conference on Computer Vision},
  pages={478--495},
  year={2024},
  organization={Springer}
}

@article{chen2021learning,
  title={Learning generalizable robotic reward functions from" in-the-wild" human videos},
  author={Chen, Annie S and Nair, Suraj and Finn, Chelsea},
  journal={arXiv preprint arXiv:2103.16817},
  year={2021}
}

@article{wu2023unleashing,
  title={Unleashing large-scale video generative pre-training for visual robot manipulation},
  author={Wu, Hongtao and Jing, Ya and Cheang, Chilam and Chen, Guangzeng and Xu, Jiafeng and Li, Xinghang and Liu, Minghuan and Li, Hang and Kong, Tao},
  journal={arXiv preprint arXiv:2312.13139},
  year={2023}
}

@inproceedings{seo2022reinforcement,
  title={Reinforcement learning with action-free pre-training from videos},
  author={Seo, Younggyo and Lee, Kimin and James, Stephen L and Abbeel, Pieter},
  booktitle={International Conference on Machine Learning},
  pages={19561--19579},
  year={2022},
  organization={PMLR}
}

@article{wu2023pre,
  title={Pre-training contextualized world models with in-the-wild videos for reinforcement learning},
  author={Wu, Jialong and Ma, Haoyu and Deng, Chaoyi and Long, Mingsheng},
  journal={Advances in Neural Information Processing Systems},
  volume={36},
  pages={39719--39743},
  year={2023}
}

@article{yang2024spatiotemporal,
  title={Spatiotemporal predictive pre-training for robotic motor control},
  author={Yang, Jiange and Liu, Bei and Fu, Jianlong and Pan, Bocheng and Wu, Gangshan and Wang, Limin},
  journal={arXiv preprint arXiv:2403.05304},
  year={2024}
}

@article{ajay2023compositional,
  title={Compositional foundation models for hierarchical planning},
  author={Ajay, Anurag and Han, Seungwook and Du, Yilun and Li, Shuang and Gupta, Abhi and Jaakkola, Tommi and Tenenbaum, Josh and Kaelbling, Leslie and Srivastava, Akash and Agrawal, Pulkit},
  journal={Advances in Neural Information Processing Systems},
  volume={36},
  pages={22304--22325},
  year={2023}
}

@article{liang2024dreamitate,
  title={Dreamitate: Real-world visuomotor policy learning via video generation},
  author={Liang, Junbang and Liu, Ruoshi and Ozguroglu, Ege and Sudhakar, Sruthi and Dave, Achal and Tokmakov, Pavel and Song, Shuran and Vondrick, Carl},
  journal={arXiv preprint arXiv:2406.16862},
  year={2024}
}

@article{hu2024video,
  title={Video prediction policy: A generalist robot policy with predictive visual representations},
  author={Hu, Yucheng and Guo, Yanjiang and Wang, Pengchao and Chen, Xiaoyu and Wang, Yen-Jen and Zhang, Jianke and Sreenath, Koushil and Lu, Chaochao and Chen, Jianyu},
  journal={arXiv preprint arXiv:2412.14803},
  year={2024}
}

@inproceedings{bruce2024genie,
  title={Genie: Generative interactive environments},
  author={Bruce, Jake and Dennis, Michael D and Edwards, Ashley and Parker-Holder, Jack and Shi, Yuge and Hughes, Edward and Lai, Matthew and Mavalankar, Aditi and Steigerwald, Richie and Apps, Chris and others},
  booktitle={Forty-first International Conference on Machine Learning},
  year={2024}
}

@article{li2025unified,
  title={Unified video action model},
  author={Li, Shuang and Gao, Yihuai and Sadigh, Dorsa and Song, Shuran},
  journal={arXiv preprint arXiv:2503.00200},
  year={2025}
}

@article{yang2023learning,
  title={Learning interactive real-world simulators},
  author={Yang, Mengjiao and Du, Yilun and Ghasemipour, Kamyar and Tompson, Jonathan and Schuurmans, Dale and Abbeel, Pieter},
  journal={arXiv preprint arXiv:2310.06114},
  volume={1},
  number={2},
  pages={6},
  year={2023}
}

@article{du2023learning,
  title={Learning universal policies via text-guided video generation},
  author={Du, Yilun and Yang, Sherry and Dai, Bo and Dai, Hanjun and Nachum, Ofir and Tenenbaum, Josh and Schuurmans, Dale and Abbeel, Pieter},
  journal={Advances in neural information processing systems},
  volume={36},
  pages={9156--9172},
  year={2023}
}

@article{zhou2024robodreamer,
  title={Robodreamer: Learning compositional world models for robot imagination},
  author={Zhou, Siyuan and Du, Yilun and Chen, Jiaben and Li, Yandong and Yeung, Dit-Yan and Gan, Chuang},
  journal={arXiv preprint arXiv:2404.12377},
  year={2024}
}

@article{rybkin2018learning,
  title={Learning what you can do before doing anything},
  author={Rybkin, Oleh and Pertsch, Karl and Derpanis, Konstantinos G and Daniilidis, Kostas and Jaegle, Andrew},
  journal={arXiv preprint arXiv:1806.09655},
  year={2018}
}

@article{mendonca2023structured,
  title={Structured world models from human videos},
  author={Mendonca, Russell and Bahl, Shikhar and Pathak, Deepak},
  journal={arXiv preprint arXiv:2308.10901},
  year={2023}
}

@article{che2024gamegen,
  title={Gamegen-x: Interactive open-world game video generation},
  author={Che, Haoxuan and He, Xuanhua and Liu, Quande and Jin, Cheng and Chen, Hao},
  journal={arXiv preprint arXiv:2411.00769},
  year={2024}
}

@article{valevski2024diffusion,
  title={Diffusion models are real-time game engines},
  author={Valevski, Dani and Leviathan, Yaniv and Arar, Moab and Fruchter, Shlomi},
  journal={arXiv preprint arXiv:2408.14837},
  year={2024}
}

@article{jang2025dreamgen,
  title={DreamGen: Unlocking Generalization in Robot Learning through Neural Trajectories},
  author={Jang, Joel and Ye, Seonghyeon and Lin, Zongyu and Xiang, Jiannan and Bjorck, Johan and Fang, Yu and Hu, Fengyuan and Huang, Spencer and Kundalia, Kaushil and Lin, Yen-Chen and others},
  journal={arXiv e-prints},
  pages={arXiv--2505},
  year={2025}
}

@article{brooks2024video,
  title={Video generation models as world simulators},
  author={Brooks, Tim and Peebles, Bill and Holmes, Connor and DePue, Will and Guo, Yufei and Jing, Li and Schnurr, David and Taylor, Joe and Luhman, Troy and Luhman, Eric and others},
  journal={OpenAI Blog},
  volume={1},
  number={8},
  pages={1},
  year={2024}
}

@article{li2024behavior1k,
    title   = {BEHAVIOR-1K: A Human-Centered, Embodied AI Benchmark with 1,000 Everyday Activities and Realistic Simulation},
    author  = {Chengshu Li and Ruohan Zhang and Josiah Wong and Cem Gokmen and Sanjana Srivastava and et al.},
    journal = {arXiv preprint arXiv:2403.09227},
    year    = {2024}
}

@inproceedings{kim2025pyroki,
  title={PyRoki: A Modular Toolkit for Robot Kinematic Optimization},
  author={Kim*, Chung Min and Yi*, Brent and Choi, Hongsuk and Ma, Yi and Goldberg, Ken and Kanazawa, Angjoo},
  booktitle={2025 IEEE/RSJ International Conference on Intelligent Robots and Systems (IROS)},
  year={2025},
  url={https://arxiv.org/abs/2505.03728},
}

@misc{liang2025zeroshotmsf,
  title={Zero-Shot Monocular Scene Flow Estimation in the Wild},
  howpublished={CVPR 2025, CVF Open Access},
  year={2025},
}

@inproceedings{wang2025skil,
  title={SKIL: Semantic Keypoint Imitation Learning for Generalizable Data-efficient Manipulation},
  author={Wang, Shengjie and You, Jiacheng and Hu, Yihang and Li, Jiongye and Gao, Yang},
  booktitle={Robotics: Science and Systems (RSS)},
  year={2025}
}

@article{yang2024depth,
  title={Depth anything v2},
  author={Yang, Lihe and Kang, Bingyi and Huang, Zilong and Zhao, Zhen and Xu, Xiaogang and Feng, Jiashi and Zhao, Hengshuang},
  journal={Advances in Neural Information Processing Systems},
  volume={37},
  pages={21875--21911},
  year={2024}
}

@inproceedings{wu2024ptv3,
    title={Point Transformer V3: Simpler, Faster, Stronger},
    author={Wu, Xiaoyang and Jiang, Li and Wang, Peng-Shuai and Liu, Zhijian and Liu, Xihui and Qiao, Yu and Ouyang, Wanli and He, Tong and Zhao, Hengshuang},
    booktitle={CVPR},
    year={2024}
}

@article{zhu2022viola,
  title={VIOLA: Imitation Learning for Vision-Based Manipulation with Object Proposal Priors},
  author={Zhu, Yifeng and Joshi, Abhishek and Stone, Peter and Zhu, Yuke},
  journal={arXiv preprint arXiv:2210.11339},
  doi={10.48550/arXiv.2210.11339},
  year={2022}
}

@MISC{coumans2021pybullet,
author =   {Erwin Coumans and Yunfei Bai},
title =    {PyBullet, a Python module for physics simulation for games, robotics and machine learning},
howpublished = {\url{http://pybullet.org}},
year = {2016--2021}
}

@inproceedings{robosuite2020,
  title={robosuite: A Modular Simulation Framework and Benchmark for Robot Learning},
  author={Yuke Zhu and Josiah Wong and Ajay Mandlekar and Roberto Mart\'{i}n-Mart\'{i}n and Abhishek Joshi and Kevin Lin and Soroush Nasiriany and Yifeng Zhu},
  booktitle={arXiv preprint arXiv:2009.12293},
  year={2020}
}

@InProceedings{haarnojaSAC2018,
  title = 	 {Soft Actor-Critic: Off-Policy Maximum Entropy Deep Reinforcement Learning with a Stochastic Actor},
  author =       {Haarnoja, Tuomas and Zhou, Aurick and Abbeel, Pieter and Levine, Sergey},
  booktitle = 	 {Proceedings of the 35th International Conference on Machine Learning},
  pages = 	 {1861--1870},
  year = 	 {2018},
  volume = 	 {80},
  series = 	 {Proceedings of Machine Learning Research},
  month = 	 {10--15 Jul},
  publisher =    {PMLR},
}

\newpage
\appendix
\subsection{Video Generation Prompts}
\label{sec:vid_gen_prompts}

 For all real world tasks, the prompt to each video generation model consists of the initial RGB observation from one camera, as well as a language instruction of the form:

\begin{promptbox}[title={Real World Task Language Prompt}]

$<$TASK$>$ by one hand. The camera holds a still pose, not zooming in or out.

\end{promptbox}

\textbf{Values of \texttt{<TASK>}:}
\begin{itemize}
    \item \textit{Put Bread in Bowl:} The bread is grabbed and placed into the green bowl
    \item \textit{Open Oven:} The toaster oven is opened
    \item \textit{Cover Bowl:} The scarf is lifted by a corner and directly dragged over the blue bowl in one smooth motion without flinging the cloth or any dynamic / fast motions
    \item \textit{Pull Out Chair:} The chair is pulled out straight from under the table to the right by grabbing the middle
    \item \textit{Open Drawer:} The partially opened drawer is opened all the way out
    \item \textit{Sweep Pasta:} The brush with a green handle moves left to right to push the pasta into the compost bin
    \item \textit{Recycle Can:} The can is grabbed and dropped into the recycling bin
\end{itemize}

For the robustness evaluations, the word ``bread'' is replaced with ``donut'' or ``long piece of bread'' for the different object instances, but otherwise the prompt remains the same.

Since \algo leverages the position of a hand in the video to help select the grasp on the object, the ``by one hand'' phrase must be included. Additionally, ``the camera holds a still pose'' must be included to increase the probability that the generated videos do not have substantial camera motion, as the depth estimation pipeline assumes a still camera. 

\begin{promptbox}[title={Push-T Task Language Prompt}]

The T-shaped block slides and rotates smoothly to the center of the wooden platform.

\end{promptbox}

\begin{figure}[!thbp]
    \centering
    \includegraphics[width=0.96\linewidth]{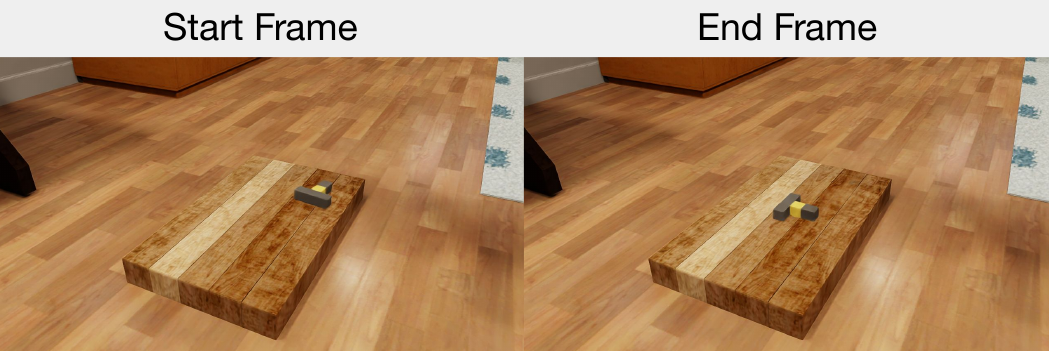}
    \caption{\textbf{Example Push-T image prompt.} To generate sufficiently accurate motions for the Push-T task, the visual component of the prompt includes both a start and end frame.}
    \label{fig:pusht_prompt}
    \vspace{-0.5em}
\end{figure}

For Push-T, since the task success requires accurate positions and to provide a better hint as to where the T block should go, we also provide a goal image of the T-block, as shown in Figure~\ref{fig:pusht_prompt}. At the time of evaluation, Veo 3 did not have the ability to take in an end frame, and hence was not considered for Push-T experiments. 

\begin{promptbox}[title={Open Door Task Language Prompt}]

The door opens all the way by itself to the right. The camera holds a still pose, not zooming in or out.

\end{promptbox}

The Open Door task prompt does not include a hand to perform the action as it is in a simulated environment. 

All prompts discussed in this section are identical for each video generation model evaluated. For Kling 2.1, a relevance value of 0.7 and a negative prompt of ``fast motion, morphing, camera motion'' are added.

\subsection{Particle Dynamics Model}
\label{sec:particle_dynamics_details}

\begin{figure}[!t]
    \centering
    \includegraphics[width=0.9\linewidth]{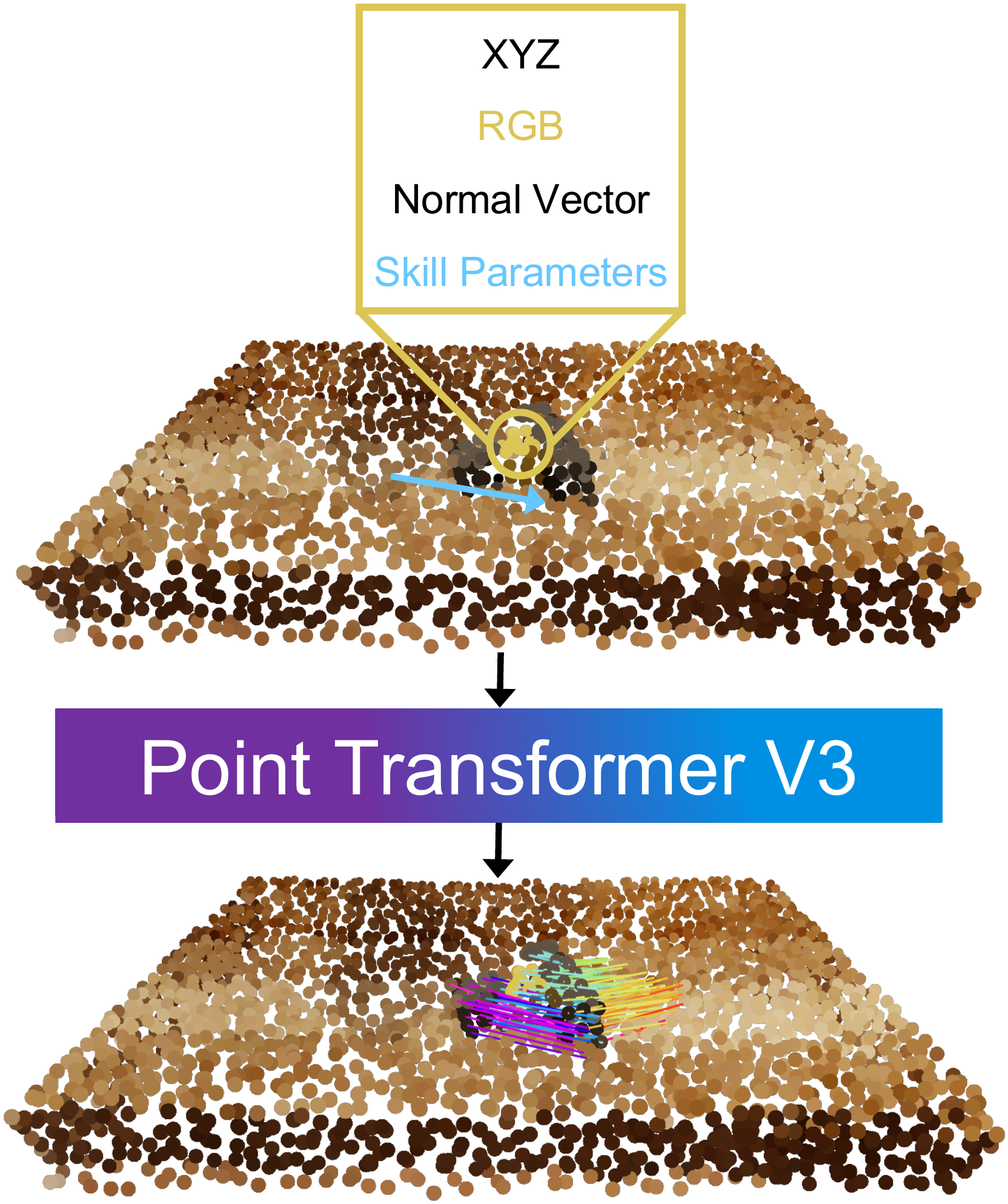}
    \caption{\textbf{Particle dynamics model.} The particle dynamics model primarily composed of a Point Transformer V3 backbone used in the Push-T task takes as input a set of feature-augmented particles and outputs delta position predictions for all particles in the scene.}
    \label{fig:particle_dynamics_model}
    \vspace{-0.5em}
\end{figure}

The particle dynamics model used in the Push-T task takes as input feature-augmented particles $\Tilde{x}_t \in \mathbb{R}^{N \times 14}$, consisting of the position, RGB value, normal vector, and push parameters and produces $\Delta \hat{x}_{t+1}$, the delta of positions of each point for the next timestep, as shown in Figure~\ref{fig:particle_dynamics_model}. The architecture of the particle dynamics model is a small Point Transformer V3~\cite{wu2024ptv3} backbone surrounded by MLPs to project between the desired input and output sizes. 

To get the input set of particles from the scene, there are 4 virtual cameras around the workspace, whose RGB-D observations are combined into a single point cloud. Then, points outside of the bounds of the wooden platform that the T-shaped block is on are discarded. Finally, this point cloud is voxel downsampled by randomly keeping only 1 particle per each 1.5cm sided cube. The same push parameter is appended to each particle's features. 

To train the particle dynamics model, we collect 500 transitions of random pushing actions. In this setting, we track particle positions before and after the push by using all objects' poses from the simulator for efficiency, but in practice this can also be tracked with CoTrackerV3~\cite{karaev2024cotracker3} and depth. 

\subsection{Push-T Planning Details}
\label{sec:pusht_planning}
To optimize the particle trajectory following cost with the learned particle based dynamics model, \algo replans after every single push with random shooting until the T-block is within the specified tolerance to the goal or a maximum number of pushes have been executed. In this setting, at each replanning step, $r$ push skill parameters are randomly sampled such that all pushes will make contact with the object of interest at different points and directions. Then, \algo selects the push skill parameter out of those $r$ ones that has the least cost according to the predicted particle positions from the dynamics model with respect to a subgoal of particle positions (may not necessarily be the final goal position). While the T-block is being pushed, \algo uses the online version of CoTrackerV3~\cite{karaev2024cotracker3} to track its motion, and once the push is completed, the tracked points are lifted into 3D, forming the updated particles of the T-block. We find that while parts of the T-block may be occluded during a pushing action, CoTrackerV3~\cite{karaev2024cotracker3} tends to recover visibility of previously occluded points when the gripper lifts up.

Since \algo tracks particles for the T-block while the particle dynamics model takes in downsampled particles of the entire scene, \algo performs nearest neighbor matching to find correspondences between the currently tracked particles and the feature-augmented particles which are the input to the dynamics model. With these correspondences, and the predicted delta positions of these corresponding particles, \algo computes the predicted particle positions of the T-block as $\hat{x}_{t+1} = \hat{x}_{t} + \Delta \hat{x}_{t+1}$.
 
For determining which timestep from the video should be used in the cost function as a subgoal, \algo first finds the timestep $t^{\star}$ where the tracked particles are closest to the particles in the \flow, as a single push can encompass the motion of many timesteps. \algo then chooses the particles from the timestep $\min(t^{\star} + L, t_{\text{end}})$ as the next subgoal for planning, where L is a fixed look ahead time amount, and $t_{\text{end}}$ is the final timestep of the video. In our experiments, we used $L=20$. We choose to use intermediate subgoals to be the target for random shooting as opposed to only the final particle positions of the T-block, as for cases involving substantial rotation, only having the final positions can result in the T-block have small translation error but large rotation error, eventually lead to timeouts.

\subsection{Grasp Selection}
\begin{figure}[!t]
    \centering
    \includegraphics[width=0.9\linewidth]{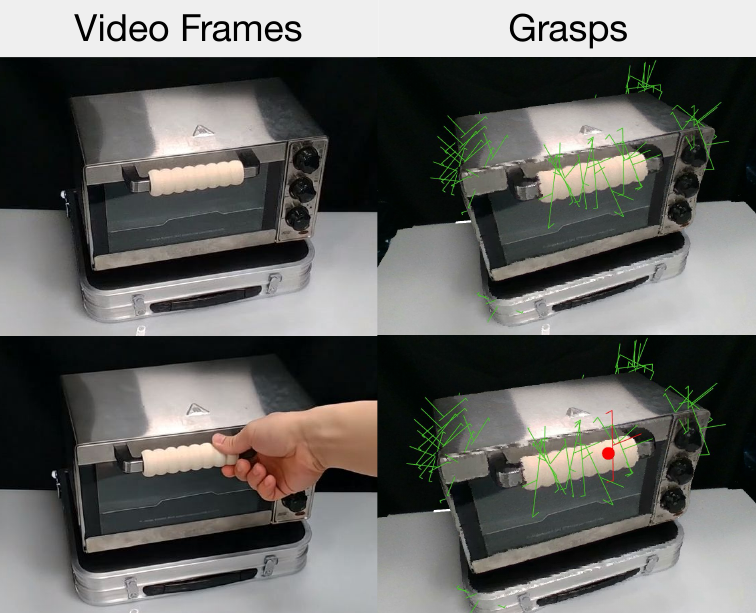}
    \caption{\textbf{Grasp selection with thumb position.} \algo selects a grasp closest to the position of the thumb in the video, if such a grasp exists within 2cm. The red sphere indicates the thumb position predicted by HaMer, with the red grasp being the selected one.}
    \label{fig:grasp_selection_thumb}
    \vspace{-0.5em}
\end{figure}

\algo uses AnyGrasp~\cite{fang2023anygrasp} to propose a set of up to 40 top-down grasps after applying the mask to the object of interest. Up to 20 grasps come from a point cloud in the coordinate frame of the camera, and another 20 grasps come from transforming those points into the coordinate frame of a virtual camera looking straight down in the center of the workspace. We added this addition virtual camera so that some more vertical grasps could be proposed. 

While for rigid objects consisting of one part, such as a piece of bread, it is typically fine to grasp at any stable position, for articulated objects, the part that moves needs to be grasped instead. To perform grasp selection in such cases, we exploit video generation models' ability to synthesize plausible hand-object interactions, typically where the hand first grabs the relevant part of the object and then proceeds to move it. \algo uses HaMer~\cite{pavlakos2024reconstructing} to detect the position of the hand, and in particular the thumb. If the predicted thumb position comes within 2cm of a proposed grasp, then such a grasp at the earliest timestep is chosen. An example of grasp selection with this method is shown in Figure~\ref{fig:grasp_selection_thumb}.

It is possible that there are no such grasps detected, either because the grasp planner proposes grasps that are not near the hand in the video or the detections from HaMer~\cite{pavlakos2024reconstructing} are not accurate enough. In such cases, \algo defaults to a heuristic of selecting the closest grasp to the points on the movable part of the rigid object (obtaining points on the movable part is described in the next section). 

\subsection{Movable Part Flow Filtering for Rigid-Grasp Dynamics Model}

\begin{figure*}[!ht]
    \centering
    \includegraphics[width=0.99\linewidth]{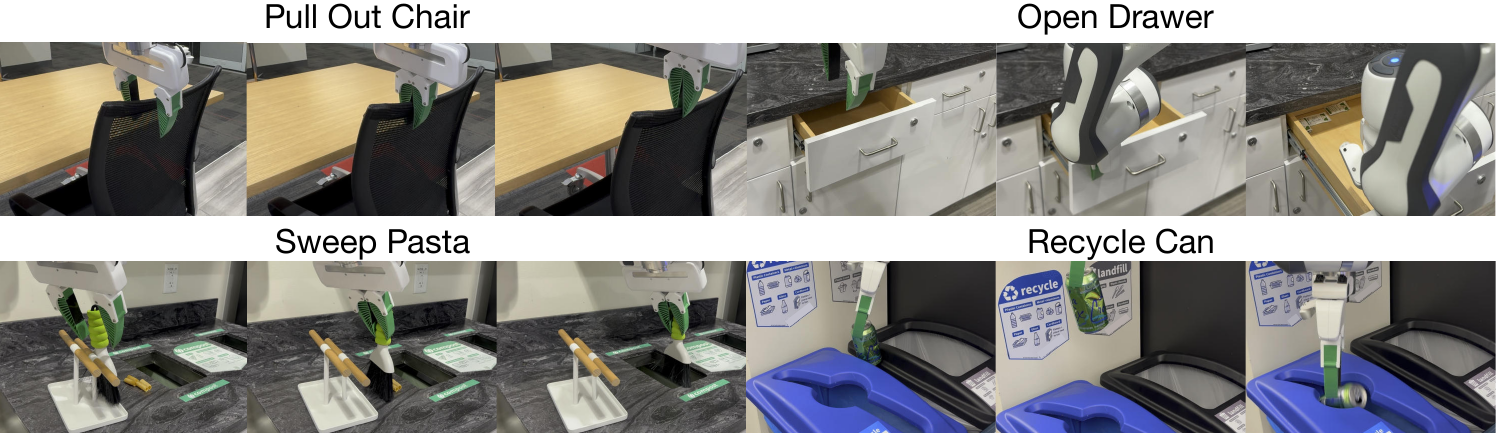}
    \caption{\textbf{In-the-Wild Task Rollouts.} The Franka successfully pulls out a chair, opens a partially opened drawer, sweeps pasta into the compost bin, and recycles an aluminum can.}
    \label{fig:in_the_wild_rollout}
    \vspace{-0.5em}
\end{figure*}

Since the rigid-grasp dynamics model assumes that points that are grasped move with the end-effector, it is necessary to identify what points are part of the movable object. To determine this, we employ the heuristic that individual flows that move at least 1 pixel on average per timestep are part of the movable flow. We empirically chose the 1 pixel threshold, as we noticed that points tracked on the stationary parts of articulated objects such as the oven had low averages of how many pixels they moved.

For the 2D tracks associated with the movable part, it may be possible that for certain intermediate frames they are 1 pixel off, making the individual tracks belong to the background or floor when lifted into 3D, causing part of the 3D object flow to be dramatically incorrect, leading to downstream execution failures (such as trying to push the toaster oven's door in the direction of the hinge). To remedy this issue, we utilize SAM 2~\cite{ravi2024sam2} with positive point prompts in the initial frame for the movable part and negative point prompts for the non-movable part, and use the part mask over each consecutive frame to constrain the 2D flow (if any tracked point is outside of the mask, it is considered to be invalid). 

\subsection{Real World Planning Details}
\label{sec:real_world_planning}

The optimization problem with a rigid-grasp assumption in the real world follows the same formulation as in Sec.~\ref{sec:problem}, where we optimize robot joint angles $\mathbf{q} \in \mathbb{R}^7$ to minimize:

\begin{equation}
\sum_{t=0}^{H-1} \lambda_{\text{task}}\big(\hat{x}^{\text{obj}}_t, \tilde{P}_t\big) + \lambda_{\text{control}}(\hat{x}_t, u_t)
\end{equation}

The control cost $\lambda_{\text{control}}(\hat{x}_t, u_t)$ is expanded into three components:
\begin{equation}
\lambda_{\text{control}}(\hat{x}_t, u_t) = w_r \mathcal{C}_r(\mathbf{q}_t) + w_s \mathcal{C}_s(\mathbf{q}_t, \mathbf{q}_{t-1}) + w_m \mathcal{C}_m(\mathbf{q}_t)
\end{equation}

where:
\begin{itemize}
\item $\mathcal{C}_r(\mathbf{q}_t)$: Reachability cost penalizing joint configurations outside the robot's workspace
\item $\mathcal{C}_s(\mathbf{q}_t, \mathbf{q}_{t-1})$: Pose smoothness cost measuring the difference between consecutive end-effector poses after running forward kinematics
\item $\mathcal{C}_m(\mathbf{q}_t)$: Manipulability cost encouraging configurations with good manipulability
\end{itemize}

The weight parameters are set to $w_r = 100$, $w_s = 1$, and $w_m = 0.01$ to encourage the robot to make smooth motions within its joint limits. The task cost $\lambda_{\text{task}}$ uses a weight of $w_f = 10$ and follows the same formulation as in Sec.~\ref{sec:problem}. The optimized end-effector poses after running forward kinematics are then fit with a B-spline and sampled such that each sampled pose is at least 1cm away from the previous and next pose and/or has a rotation difference of at least 20 degrees. To execute this planned trajectory, we use the IK solver from PyBullet~\cite{coumans2021pybullet} to get target joint angles and then a joint impedance controller from Deoxys~\cite{zhu2022viola} commands the Franka to reach those positions. 

\subsection{In-the-Wild Tasks}
\label{sec:in_the_wild_desc}

For In-the-Wild tasks, we consider:

\subsubsection{Pull Out Chair} 
In the Pull Out Chair task, a black rolling chair is placed underneath a table. 
A trial is considered a success if the chair is moved at least 5cm horizontally from its initial position. The reason for the limited motion requirement is that in certain configurations, it is difficult for the Franka robot to push the chair. 

\subsubsection{Open Drawer}
The Open Drawer task involves opening a partially opened drawer out to at least 90\% of its full possible extension. 

\subsubsection{Sweep Pasta}
In the Sweep Pasta task, there is a brush with a green handle placed against a wooden support structure, along with four pieces of dried pasta next to a compost bin. The objective is for the robot to grasp the handle of the brush and use the brush to push the pasta into the compost bin. If all pieces of pasta are inside of the compost bin, it is considered a success.  

\subsubsection{Recycle Can}
In the Recycle Can task, an aluminum can is placed in between a recycling and trash bin while upright. A trial is successful if the can is inside of the recycling bin.

See Fig.~\ref{fig:in_the_wild_rollout} for rollouts for each of these in-the-wild tasks.

\subsection{Open Door Reinforcement Learning Details}
\label{sec:open_door_rl}

For the Open Door task, we utilize Soft Actor-Critic (SAC)~\cite{haarnojaSAC2018} to train sensorimotor policies across three different embodiments: a Franka Panda, a Spot robot, and a GR1 humanoid arm. The policies are trained to follow the 3D object flow extracted from generated videos, which serves as a reward signal. 

\subsubsection{Hyperparameters}
The hyperparameters used for SAC training are detailed in Table~\ref{tab:sac_hyperparams}. These values were consistent across all reward types and embodiments with exception of the GR1, which uses 10000 training iterations due to the larger action space.

\begin{table}[h]
    \centering
    \begin{tabular}{lc}
        \toprule
        Hyperparameter & Value \\
        \midrule
        Learning rate & $3 \times 10^{-4}$ \\
        Discount factor ($\gamma$) & 0.99 \\
        Batch size & 256 \\
        Buffer size & $10^6$ \\
        Target update rate ($\tau$) & 0.005 \\
        Target network update freq & 1 \\
        Hidden layers & 2 \\
        Hidden units per layer & 256 \\
        Training iterations & 5000 \\
        Episode horizon & 500 \\
        \bottomrule
    \end{tabular}
    \caption{\textbf{SAC Hyperparameters for Open Door Task.}}
    \label{tab:sac_hyperparams}
\end{table}

\subsubsection{Reward Functions}
We compare policies trained with two different reward formulations: a handcrafted object state reward and a 3D object flow reward.

\textbf{Object State Reward:} The handcrafted reward $R_{\text{state}}$ consists of a reaching component $r_{\text{reach}}$ and a handle rotation component $r_{\text{rot}}$:
\begin{equation}
    r_{\text{reach}} = 0.25 \cdot (1 - \tanh(10 \cdot d_{\text{gripper, handle}}))
\end{equation}
\begin{equation}
    r_{\text{rot}} = \text{clip}\left(0.25 \cdot \frac{|\theta_{\text{handle}}|}{0.5\pi}, -0.25, 0.25\right)
\end{equation}
where $d_{\text{gripper, handle}}$ is the L2 distance between the robot's end-effector and the door handle, and $\theta_{\text{handle}}$ is the rotation angle of the handle. The total reward is $r_{\text{reach}} + r_{\text{rot}}$, unless the door hinge angle $\theta_{\text{hinge}} > 0.3$ radians, in which case a completion reward of 1.0 is returned.

\textbf{3D Object Flow Reward:} The 3D object flow reward $R_{\text{flow}}$ leverages the reference trajectory $P_{1:T}$ extracted from the video. It consists of a particle tracking term $r_{\text{particle}}$ and an end-effector alignment term $r_{\text{ee}}$:
\begin{equation}
    r_{\text{particle}} = 0.75 \cdot \frac{t^\star}{t_{\text{end}}}
\end{equation}
where $t^\star$ is the index of the closest timestep in the reference trajectory $P_{1:T}$ based on the current object particle positions $\hat{x}_t^{\text{obj}}$ in the door frame:
\begin{equation}
    t^\star = \argmin_{t \in \{1 \dots t_{\text{end}}\}} \frac{1}{n} \sum_{i=1}^n \|\hat{x}_t^{\text{obj}}[i] - P_t[i]\|_2
\end{equation}
The end-effector term $r_{\text{ee}}$ encourages the robot to stay near the object particles:
\begin{equation}
    r_{\text{ee}} = 0.25 \cdot (1 - \tanh(10 \cdot \|ee_{\text{door}} - \bar{x}\|_2))
\end{equation}
where $ee_{\text{door}}$ is the end-effector position in the door body frame and $\bar{x} = \frac{1}{n}\sum_{i=1}^n \hat{x}_t^{\text{obj}}[i]$ is the mean position of the object particles. The total reward is $R_{\text{flow}} = r_{\text{particle}} + r_{\text{ee}}$.

Instead of tracking the mean particle position with CoTrackerV3, we proceed to use a simpler approach of transforming the initial particles as the door angle changes for better computation efficiency. We thus consider the door joint angle to be a part of the state $x_t$.

\subsection{Video Generation Failures}
\begin{figure}[!thbp]
    \centering
    \includegraphics[width=0.96\linewidth]{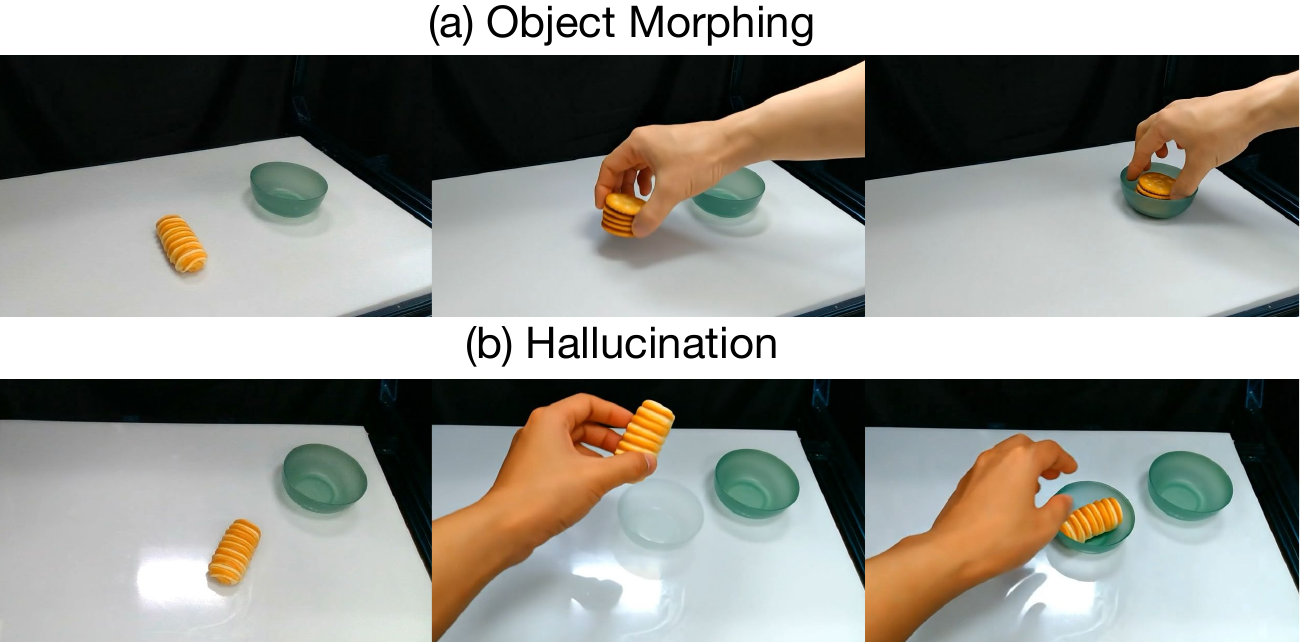}
    \caption{\textbf{Video Generation Failure Examples.} (a) The bread undergoes object morphing, where it turns into a stack of crackers, causing the downstream tracking to fail. (b) Another green bowl appears due to model hallucination, causing a downstream execution failure where the bread is dropped onto the surface of the workspace instead of the original green bowl.}
    \label{fig:vid_gen_failures}
    \vspace{-0.5em}
\end{figure}

From the generated videos, we observe that there are two common failure modes: morphing and hallucination. Object morphing occurs when an existing object in the scene dramatically changes shape to something else, such as another object or an object with significantly different geometric properties than what it should be. Hallucination occurs when a new object (previously non-existent) appears in the scene. We show examples of morphing and hallucination for the Put Bread task in Fig.~\ref{fig:vid_gen_failures}.

\subsection{Limitations}
\algo has several limitations. First, it relies on a rigid-grasp assumption for real world manipulation, limiting the types of tasks that can be performed. While this work shows that a particle dynamics model can be used for other types of tasks such as non-prehensile pushing, training and scaling a particle dynamics model for the real world is non-trivial and can be considered for future work. Another limitation is that the total processing time to get 3D object flow depending on the video generation model is between 3 and 11 minutes, which limits its usability, with the main bottleneck being the video generation. Since \algo relies upon one angle in a generated video, it cannot handle heavy occlusions gracefully, such as when the human hand covers the majority of a small object. Future work may consider methods which can deal with such occlusions better, such as 3D point trackers~\cite{xiao2025spatialtracker} or full 4D representations.

\end{document}